%% file: main_camera_ready.tex
\documentclass[conference]{IEEEtran}
\IEEEoverridecommandlockouts
\usepackage{cite}
\usepackage{amsmath,amssymb,amsfonts}
\usepackage{algorithmic}
\usepackage{graphicx}
\usepackage{textcomp}
\usepackage{xcolor}
\def\BibTeX{{\rm B\kern-.05em{\sc i\kern-.025em b}\kern-.08em
    T\kern-.1667em\lower.7ex\hbox{E}\kern-.125emX}}

\usepackage{hyperref}
\usepackage{enumitem}
\usepackage{styles/math_commands}

\input{contents/shared_macros}

\makeatletter
\let\MYcaption\@makecaption
\makeatother

\usepackage[font=footnotesize]{subcaption}

\makeatletter
\let\@makecaption\MYcaption
\makeatother

\begin{document}

\input{contents/title}

\input{contents/authors}

\maketitle

\input{contents/abstract_keywords}

\input{contents/main}

{
\bibliography{ref}
}

\input{contents/appendix}

\end{document}

%% file: contents/shared_macros.tex
\usepackage{amsmath,amsfonts,amsthm,commath}
\usepackage{microtype}
\usepackage{graphicx,xcolor}
\usepackage{booktabs,multirow,multicol}
\PassOptionsToPackage{hyphens}{url}
\usepackage{hyperref}
\usepackage{cleveref}
\usepackage{siunitx}
\usepackage{tikz,tikz-qtree}
\usepackage{bbding} 
\usepackage{lipsum} 
\usepackage{xurl}
\usepackage{colortbl}
\usepackage{float}
\usepackage[most]{tcolorbox}

\usepackage{styles/math_commands}

\definecolor{tabgray}{gray}{0.90}
\definecolor{tlegray}{gray}{0.5}

\newcount\Comments  
\newcount\arxiv 
\newcount\CSUR 
\newcount\JMLR 
\newcount\IEEE 
\usepackage{color}
\definecolor{darkgreen}{rgb}{0,0.5,0}
\definecolor{darkblue}{rgb}{0,0,0.8}
\definecolor{orange}{rgb}{0.96,0.4,0.25}
\definecolor{purple}{rgb}{1,0,1}
\definecolor{rev}{rgb}{0,0,0}
\newcommand{\kibitz}[2]{\ifnum\Comments=0\textcolor{#1}{#2}\fi}

\ifnum\Comments=0
\fi

\definecolor{rev}{rgb}{0,0,0.9}

\newcommand{\cL}{\ell}

\newcommand{\cA}{\gA}
\newcommand{\cM}{\gM}
\newcommand{\cD}{\gD}
\newcommand{\cX}{\gX}
\newcommand{\cS}{\gS}
\newcommand{\bW}{\mW}
\newcommand{\bL}{\mL}
\newcommand{\bU}{\mU}
\newcommand{\bP}{\mP}
\newcommand{\bbR}{\sR}

\newcommand{\hz}{\hat{z}}

\newcommand{\poly}{\mathrm{poly}}
\newcommand{\Std}{\mathrm{Std}}

\newcommand{\para}[1]{{\vspace{2pt} \bf \noindent #1 \hspace{3pt}}}

\renewcommand{\sc}[1]{\textsc{#1}}
\newcommand{\mc}[2]{\multicolumn{#1}{c|}{#2}}
\newcommand{\emc}[2]{\multicolumn{#1}{c}{#2}}
\newcommand{\mb}[1]{\mathbf{#1}}

\renewcommand{\t}{\textsc}
\newcommand{\tle}[1]{\textcolor{tlegray}{#1}}

\newcommand{\cellgray}{\cellcolor{tabgray}}

\Crefname{definition}{Def.}{Defs.}
\Crefname{section}{Sec.}{Secs.}
\Crefname{figure}{Fig.}{Fig.}
\Crefname{equation}{Eqn.}{Eqns.}

\newenvironment{practitioner}
  {
	\begin{tcolorbox}
	[%
    enhanced, 
    breakable,
    boxrule=0.1pt,
    left=1pt,
    right=1pt,
    bottom=1pt,
    top=1pt
    ]{}
  \textbf{Practical implications.}
  \small}
  {
\end{tcolorbox}
}

\newenvironment{researcher}
  {
    \vspace{-0.2em}
	\begin{tcolorbox}
	[%
    enhanced, 
    breakable,
    boxrule=0.1pt,
    left=1pt,
    right=1pt,
    bottom=1pt,
    top=1pt
    ]{}
  \textbf{Research implications.}
  \small}
  {
\end{tcolorbox}
    \vspace{-0.6em}
}

%% file: contents/title.tex
\title{\texorpdfstring{SoK: Certified Robustness for\\ Deep Neural Networks}{SoK: Certified Robustness for Deep Neural Networks}}

%% file: contents/authors.tex
\ifnum\CSUR=1

\author{Linyi Li}
\email{linyi2@illinois.edu}
\affiliation{%
  \institution{University of Illinois Urbana-Champaign}
  \streetaddress{201 N. Goodwin Ave}
  \city{Urbana}
  \state{Illinois}
  \country{USA}
  \postcode{61801}
}

\author{Tao Xie}
\email{taoxie@pku.edu.cn}
\affiliation{%
  \institution{Key Laboratory of High Confidence Software Technologies, MoE (Peking University)}
  \streetaddress{5 Yiheyuan Rd}
  \city{Beijing}
  \country{China}
  \postcode{100871}
}

\author{Bo Li}
\email{lbo@illinois.edu}
\affiliation{%
  \institution{University of Illinois Urbana-Champaign}
  \streetaddress{201 N. Goodwin Ave}
  \city{Urbana}
  \state{Illinois}
  \country{USA}
  \postcode{61801}
}

\fi

\ifnum\JMLR=1

\author{\name Linyi Li \email linyi2@illinois.edu \\
        \addr Department of Computer Science\\
        University of Illinois Urbana Champaign\\
        Urbana, IL 61801, USA
        \AND
        \name Tao Xie \email taoxie@pku.edu.cn \\
        \addr Key Laboratory of High Confidence Software Technologies, MoE\\
        Peking University\\
        Beijing 100871, China
        \AND
        \name Bo Li \email lbo@illinois.edu \\
        \addr Department of Computer Science\\
        University of Illinois Urbana Champaign\\
        Urbana, IL 61801, USA
}

\editor{Francis Bach and David Blei and Bernhard Sch{\"o}lkopf}
\fi

\ifnum\IEEE=1

    



\author{
    \IEEEauthorblockN{
        Linyi Li\IEEEauthorrefmark{1} 
        \quad
        Tao Xie\IEEEauthorrefmark{2}
        \quad
        Bo Li\IEEEauthorrefmark{1}
    }
    
  \IEEEauthorblockA{\IEEEauthorrefmark{1} University of Illinois Urbana-Champaign, \{\href{mailto:linyi2@illinois.edu}{linyi2},\href{mailto:lbo@illinois.edu}{lbo}\}@illinois.edu}
  \IEEEauthorblockA{\IEEEauthorrefmark{2} Key Laboratory of High Confidence Software Technologies, MoE (Peking University), \href{mailto:taoxie@pku.edu.cn}{taoxie@pku.edu.cn}}

  \IEEEauthorblockA{Version 9 (Apr 12, 2023)}
}

\fi

%% file: contents/abstract_keywords.tex
\begin{abstract}%
Great advances in deep neural networks~(DNNs) have led to state-of-the-art performance on a wide range of tasks. However, recent studies have shown that DNNs are vulnerable to adversarial attacks, which have brought great concerns when deploying these models to safety-critical applications such as autonomous driving.
Different defense approaches have been proposed against adversarial attacks, including: a) \textit{empirical defenses}, which can usually be adaptively attacked again without providing robustness certification; and b)
\textit{certifiably robust approaches}, which consist of \textit{robustness verification} providing the lower bound of robust accuracy against \textit{any} attacks under certain conditions and corresponding \textit{robust training} approaches.
In this paper, we systematize certifiably robust approaches and related practical and theoretical implications and findings. We also provide the \textit{first} comprehensive benchmark on existing robustness verification and training approaches on different datasets. In particular, we 
1)~provide a taxonomy for the robustness verification and training approaches, as well as summarize the methodologies for representative algorithms, 
2)~reveal the characteristics, strengths, limitations, and fundamental connections among these approaches, 
3)~discuss current research progresses, theoretical barriers, main challenges, and future directions for certifiably robust approaches for DNNs, 
and 4)~provide an open-sourced unified platform to evaluate 20+ representative certifiably robust approaches.
\end{abstract}

\ifnum\CSUR=1

\ccsdesc[500]{General and reference~Surveys and overviews}
\ccsdesc[500]{Computing methodologies~Neural networks}
\ccsdesc[300]{Software and its engineering~Formal software verification}

\keywords{neural networks, robustness certification, adversarial examples, verification, survey}

\fi

\ifnum\JMLR=1


\fi

\ifnum\IEEE=1


\fi

%% file: contents/main.tex
\section{Introduction}
    \label{sec:intro}

    Machine learning~(ML) techniques, especially deep neural networks~(DNNs), have been widely adopted in various applications, such as image classification~\cite{he2016deep,krizhevsky2009learning,szegedy2016rethinking} and natural language processing~\cite{brown2020language,devlin2018bert,vaswani2017attention}.
    However, despite their wide applications, 
    both traditional ML models~\cite{biggio2013evasion,dalvi2004adversarial,lowd2005adversarial} and DNNs~\cite{goodfellow2014explaining,szegedy2013intriguing} are shown vulnerable to adversarial evasion attacks where carefully crafted \textit{adversarial examples} ---  inputs with adversarial perturbations --- could mislead ML models to make arbitrarily incorrect predictions~\cite{athalye2018obfuscated,brendel2017decision}.
    The existence of adversarial attacks leads to great safety concerns for DNN-based applications, especially in safety-critical scenarios such as autonomous driving~\cite{cao2021invisible,eykholt2018robust}.
    
    To defend against such attacks, there are several works proposed to empirically improve the robustness of DNNs~\cite{buckman2018thermometer,dalvi2004adversarial,guo2018countering,madry2017towards,papernot2016distillation,samangouei2018defensegan}.
    However, many of such defenses can be adaptively attacked again by sophisticated attackers~\cite{athalye2018obfuscated,tramer2020adaptive}.
    The everlasting competition between attackers and defenders motivates studies on the \textbf{certifiably robust approaches} for DNNs, which include both \textbf{robustness verification} and \textbf{robust training} approaches~\cite{cohen2019certified,katz2017reluplex,singh2019beyond,singh2018robustness,wang2018efficient,wong2018provable,yang2020randomized}.
    The robustness verification approaches aim to 
    evaluate DNN robustness  by providing a theoretically certified lower bound of robustness under certain perturbation constraints;
    the corresponding robust training approaches aim to train DNNs to improve such lower bound.

    In this paper, we aim to provide a taxonomy for existing certifiably robust approaches (i.e., robustness verification and robust training approaches) from the first principle, as well as a comprehensive benchmark on different datasets and models to enable the quantitative comparison for the community. 
    Existing surveys discuss general attacks and defenses for traditional ML models~\cite{biggio2018wild,chakraborty2018adversarial,huang2011adversarial,papernot2018sok} and DNNs~\cite{albarghouthi2021introduction,huang2020survey,liu2019algorithms,miller2020adversarial}, but they mainly focus on empirical defenses without guarantees or some specific verification approaches.
    To the best of our knowledge, this is the \textit{first} systematic taxonomy for the fast-developing \emph{certifiably robust} approaches on DNNs against \emph{evasion} attacks.
    The taxonomy reveals characteristics, strengths, limitations, and fundamental connections among these approaches.
    

    To provide quantitative analysis for existing certifiably robust approaches, we develop an open-source unified toolbox for representative verification and training approaches.
    We benchmark over 20 verification and robust training approaches.
    As far as we know, it is the \emph{first} large-scale benchmark for the certified robustness of DNNs.
    Based on the taxonomy, analysis, and benchmark of existing approaches, we further provide discussion and analysis on current research progresses, theoretical barriers, and several promising future directions.
    We also outline how to extend these approaches to alternative threat models and system models along with their applications.

    This SoK is intended for both ML experts, who aim to develop and improve certifiably robust ML approaches, as well as practical users with a focus on applying certifiably robust approaches to different real-world ML applications. 
    For ML experts, this SoK provides (1)~systematic taxonomy to contextualize their  works, 
    (2)~detailed explanation and analysis/comparison for representative certifiably robust approaches, 
    and (3)~discussion of research implications, including limitations, challenges, and future directions.
    For practical users, the SoK provides
    (1)~formal problem definition of robustness verification,
    (2)~comprehensive benchmark and reference implementations of representative approaches to ease the deployment,
    and (3)~practical implications on how to select the most suitable defenses and how to evaluate existing defenses using certifiably robust approaches.

    \begin{figure*}[t]
        \centering
         \begin{subfigure}{.165\linewidth}
             \centering
             \includegraphics[width=\linewidth]{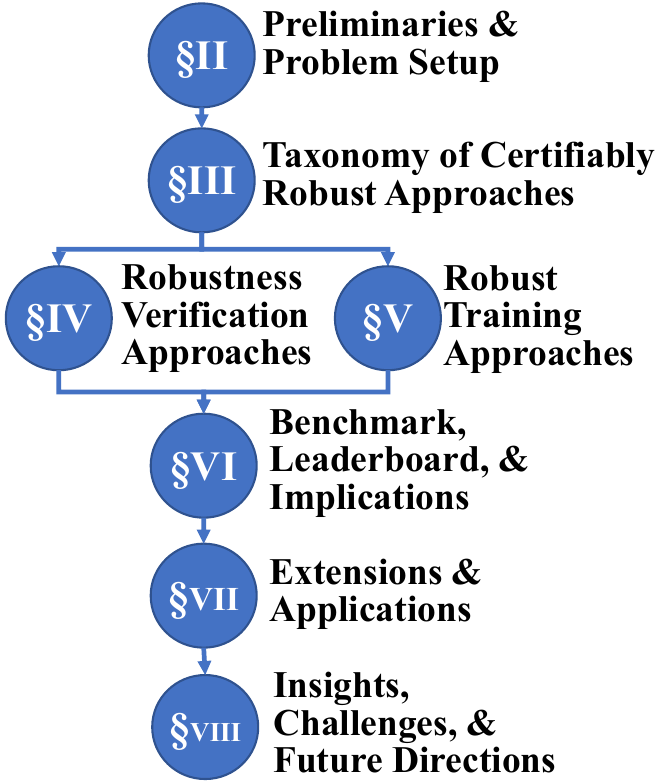}
             \vspace{-1.65em}
             \caption{\,}
             \label{fig:structure}
         \end{subfigure}
         \hspace{.03\linewidth}
         \begin{subfigure}{.65\linewidth}
             \centering
             \includegraphics[width=\linewidth]{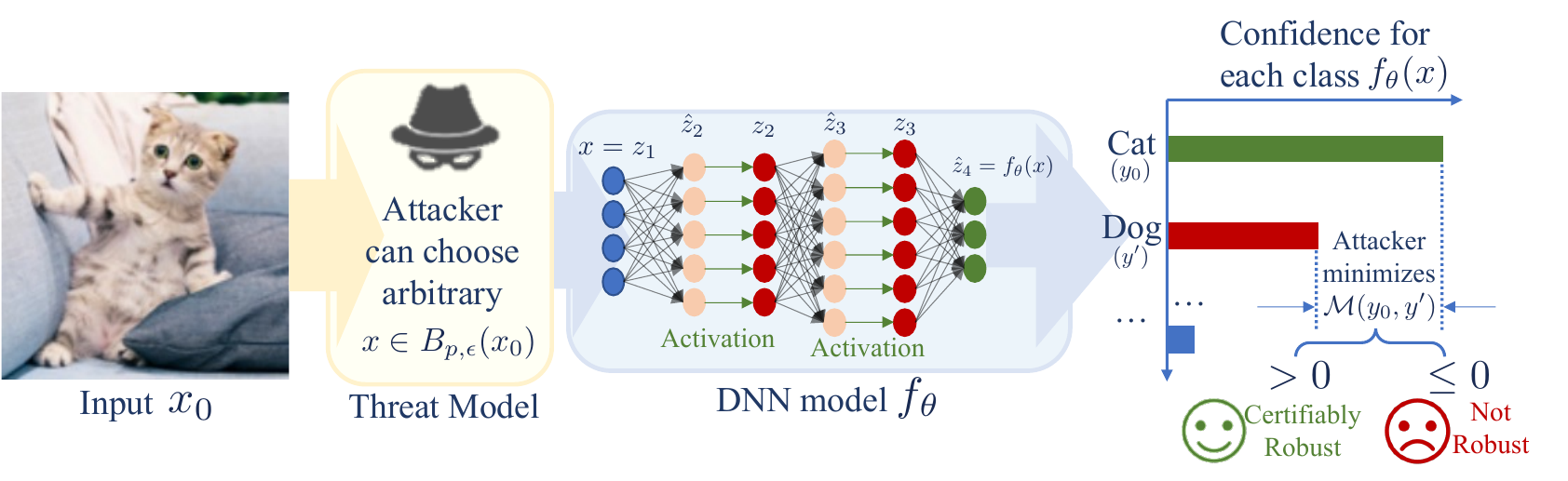}
             \vspace{-1.85em}
             \caption{\,}
             \label{fig:verification-problem}
         \end{subfigure}
        \vspace{-0.6em}
        \caption{ \textbf{(a)}~Overview of this SoK; \textbf{(b)}~Illustration of robustness verification as optimization~(\Cref{prob:verification-optimization-problem}).}
        \label{fig:overview-and-vertification-problem}
        \vspace{-1.6em}
    \end{figure*}

    In taxonomizing and analyzing certifiably robust approaches for DNNs,  we make the following \underline{contributions}:
    \begin{itemize}[leftmargin=*]
        \item  We provide a general problem definition for the robustness verification problem and the \textit{first} systematic taxonomy of certifiably robust approaches for DNNs  (\Cref{sec:main-taxonomy}), including the robustness verification approaches (\Cref{sec:verification-approaches}), and robust training approaches~(\Cref{sec:training-approach}). 
        
        \item 
        We conduct extensive quantitative comparisons\footnotemark for different state-of-the-art approaches on robustness verification and robust training, leading to a benchmark and leaderboard, from which we summarize practical implications for deploying certifiably robust approaches~(\Cref{sec:evaluation}).
        
        
    \footnotetext{The benchmark website with open-source toolbox, including full results are available at \textit{\url{https://sokcertifiedrobustness.github.io}}\ifnum\arxiv=0~(anonymized)\fi.}

        \item We provide an open-source unified evaluation toolbox for over $20$ verification and training approaches, which we believe will facilitate the development and evaluation of research on certified robustness for DNNs.\footnotemark[\value{footnote}]
        
         \item We discuss and analyze current research progresses, theoretical barriers,  challenges, extensions, and further provide several potential future research directions  (\Cref{sec:extensions,sec:discussion}).
    \end{itemize}

\section{Preliminaries and Problem Setup}
    
    In this section, we provide the preliminaries and a general problem definition for robustness verification.
    We denote $[n]$ as set $\{1,\,2,\,\dots,\,n\}$.
    To represent the region of adversarially perturbed input,
    when measured by $\cL_p$ norm~$(p \in \sN_+ \cup \{+\infty\})$ we use $B_{p,\epsilon}(x_0)$ to denote the perturbed input which is drawn from the region centered at $x_0$ with $\epsilon$ radius, i.e., $B_{p,\epsilon}(x_0) := \{x: \|x - x_0\|_p < \epsilon\}$, where $\epsilon$ is called perturbation radius.
    
    \subsection{System Model}

        We focus on the certified robustness of DNNs for classification tasks for brevity and ease of exposition. Extensions to other system models and other tasks are discussed in \Cref{sec:extensions}.
        
        \label{subsec:target-model}
        
        A (classification) DNN model $f_\theta$ is formulated as a function: $\cX \to \bbR^C$, where
        the input data $x \sim \cD$ is in a bounded $n$-dimensional subspace $\cX \subseteq [0,\,1]^n$, and the model provides confidence scores for all $C$ classes.
        $F_\theta(x) := \argmax_{i \in [C]} f_\theta(x)_i$ is the predicted class of model $f_\theta$ given input $x$.
        $\theta$ is the set of trainable parameters for $f_\theta$.
        For brevity, we may omit $\theta$ when there is no ambiguity.
        There are many different DNN architectures.
        One common system model is feed-forward ReLU networks as defined in \Cref{def:feed-forward-nn}.
        \begin{definition}[Feed-Forward ReLU Networks]
            An $l$-layer feed-forward ReLU network $f_\theta$ is defined as such:
            \vspace{-0.7em}
            \begin{equation}
                \small
                \left\{
                \begin{aligned}
                    & z_1 := x, \\
                    & \hz_{i+1} := \bW_i z_i + b_i, \quad & \text{for $i = 1,\,2,\,\dots,l-1$} \\ 
                    & z_{i} := \relu(\hz_{i}), \quad & \text{for $i = 2,\,\dots,l$} \\
                    & f_\theta(x) := \hz_l,
                \end{aligned}
                \right.
            \vspace{-0.5em}
            \end{equation}
            where 
            $\relu(z,\,0) = \max\{z, 0\}$.
            Each $z_i$ and $\hz_i$ is a vector in $\bbR^{n_i}$.
            In particular, $n_1 = n$ and $n_l = C$.
            The trainable parameters $\theta := \{ \bW_i, b_i:\, i \in [l-1] \}$.
            \label{def:feed-forward-nn}
        \end{definition}
        \noindent In \Cref{tab:property-table}, the ``System Model'' column lists some other system models which we will define when illustrating their corresponding verification approaches in \Cref{sec:verification-approaches}. 
    \subsection{Threat Model}
        \label{subsec:threat-model}
        Existing studies on certified robustness~\cite{cohen2019certified,madry2017towards,gehr2018ai2,wong2018provable,zhang2018efficient} mainly aim to defend against \textit{white-box \textbf{evasion}} attacks, which indicate the strongest adversaries who have full knowledge of the target model, including its parameters and architecture.
        In particular, the adversary would carefully craft a bounded perturbation  to the original input, generating an  \emph{adversarial example}~\cite{szegedy2013intriguing} to fool the model into making incorrect predictions.
        Formally, we define \emph{$(\cL_p,\,\epsilon)$-adversary}.
        \begin{definition}[$(\cL_p,\,\epsilon)$-Adversary]
            For given input $(x_0, y_0)$, where $x_0\in \cX$ is the input instance and $y_0 \in [C]$ is its true label, 
            the \emph{$(\cL_p,\,\epsilon)$-adversary} will generate a perturbed input $x \in B_{p,\epsilon}(x_0)$, such that $F_\theta (x) \neq y_0$.
            When there is no ambiguity, we will call it \emph{$\cL_p$ adversary}.
            \label{def:lp-adversary}
        \end{definition}
        We focus on certifiably robust approaches against $(\cL_p,\epsilon)$-adversary, since approaches for this adversary are well-developed, and approaches for other threat models can be extended from those for $(\cL_p,\epsilon)$-adversary.
        We will discuss other threat models in \Cref{sec:extensions}.
        The above definition conforms to untargeted attack whose goal is to deviate the model prediction from the ground truth.
        The targeted attack which aims to mislead the model to output a specific label $y'$, can be defined similarly.
        The literature also refers to an $(\cL_p,\,\epsilon)$-adversary as an \emph{$\cL_p$-bounded attack}~(bounded by $\epsilon$).
        To the best of our knowledge,
        existing $\cL_p$-bounded attacks only consider $p=1,2,\infty$.
        The inputs generated by these adversaries are within $\epsilon$ distance to clean input $x_0$ measured by $\ell_1$ norm~(i.e., Matthattan distance), $\ell_2$ norm~(i.e., Euclidean distance), and $\ell_\infty$ norm~(i.e., maximum difference among all dimensions) respectively.
        We illustrate the region from which the attacker picks the perturbed input in \Cref{fig:different-norms} in App.~\ref{adxsubsec:lp-adversary-fig}.
        For 2D input, the region shapes are diamond, circle, and square for $\ell_1$, $\ell_2$, and $\ell_\infty$ adversaries respectively.


    
    \subsection{Robustness Verification and Robust Training}
        \label{sec:definition-robust-verification-provable-defense}
        
        \para{Robustness verification.}
        A \emph{robustness verification} approach certifies the lower bound of model's performance against \emph{any} adversary under certain constraints, e.g., $\cL_\infty$-bounded attack. We can categorize the verification approaches into \emph{complete verification} and \emph{incomplete verification}.
        When the verification approach outputs ``not verified'' for a given $x_0$, if it is guaranteed that an adversarial example $x$ around $x_0$ exists we call it \emph{complete verification}; and otherwise \emph{incomplete verification}.
        
        We can also categorize the verification approaches into \emph{deterministic verification} and \emph{probabilistic verification}.
        When the given input is non-robust against the attack, deterministic verification is guaranteed to output ``not verified'';
        and the probabilistic verification is guaranteed to output ``not verified'' with a certain probability (e.g., $99.9\%$) where the randomness is independent of the input. 
        Formal definitions are as follows. 
        \begin{definition}[Robustness Verification]
            An algorithm $\cA$ is called a \emph{robustness verification}, if for any $(x_0, y_0)$, as long as there exists $x \in B_{p,\epsilon}(x_0)$ with $F_\theta(x) \neq y_0$~(adversarial example), $\cA(f_\theta,\,x_0,\,y_0,\,\epsilon) = \mathrm{false}$~(\emph{deterministic verification}) or $\Pr[\cA(f_\theta,\,x_0,\,y_0,\,\epsilon) = \mathrm{false}] \ge 1 - \alpha$~(\emph{probabilistic verification}), where $\alpha$ is a pre-defined small threshold.
            If $\gA(f_\theta, x_0, y_0, \epsilon) = \mathrm{true}$, we call $\gA$ provides \textbf{robustness certification} for model $f_\theta$ on $(x_0,y_0)$ against $(\ell_p,\epsilon)$-adversary.
             Whenever $\cA(f_\theta,\,x_0,\,y_0,\,\epsilon) = \mathrm{false}$, 
             if there exists $x\in B_{p,\epsilon}(x_0)$ with $F_\theta(x) \neq y_0$,  $\cA$ is called \textbf{complete verification}, otherwise \textbf{incomplete verification}.
            \label{def:verification-approach}
            \label{def:complete-verification-approach}
        \end{definition}
        If we view ``certifying a truly robust instance'' as the true positive, then a robustness verification approach produces false positives with small~(probabilistic) or zero~(deterministic) probability, and complete verification produces no false negatives.
        If the verification cannot certify an instance, it is possible that either the instance is not robust or the verification approach is too loose to certify it.
        We can also view robustness verification from optimization perspective.
        
        \begin{problem}[Robustness Verification as Optimization]
            \label{prob:verification-optimization-problem}
            Given a neural network $f_\theta: \cX \to \bbR^C$, input instance $x_0 \in \cX$, ground-truth label $y_0 \in [C]$, any other label $y' \in [C]$ and the radius $\epsilon > 0$, we define the following optimization problem:
            \begin{equation*}
                \small
                    \cM(y_0,y') = \minimize_{x}\, f_\theta(x)_{y_0} - f_\theta(x)_{y'}
                    \, 
                    \text{ s.t. }
                    \, 
                    x \in B_{p,\epsilon}(x_0).
            \end{equation*}
            If $\cM(y_0,y') > 0, \forall y'\in[C] \setminus \{y_0\}$,  $f_\theta$ is certifiably robust at $x_0$ within radius $\epsilon$ w.r.t. $\cL_p$ norm.
        \end{problem}

        Intuitively, \Cref{prob:verification-optimization-problem} searches for the minimum margin between the model confidence for the true class $y_0$ and any other class $y'$.
        For any $y' \neq y_0$, if we can certify $\gM(y_0,y')>0$,  the margin $f_\theta(x)_{y_0} - f_\theta(x)_{y'}$ is always positive. 
        Since the model will predict the class with the highest confidence, this means for any possible perturbed input $x\in B_{p,\epsilon}(x_0)$, the predicted class is always $y_0$, and therefore the robustness is certified.
        \Cref{fig:verification-problem} illustrates this process.

        The robustness verification then boils down to deciding whether $\gM(y_0, y') > 0$.
        If a procedure exactly solves \Cref{prob:verification-optimization-problem}, the corresponding verification approach is complete. 
        If a procedure conservatively provides a lower bound of $\cM$, the corresponding verification approach is usually incomplete.
        
        Although complete verification sounds attractive, it is \NP-Complete~\cite{katz2017reluplex,weng2018towards}.
        This intrinsic barrier, which we identify as \textbf{scalability challenge}, impedes complete verification approaches from scaling up to common DNN sizes. 
        To overcome this scalability challenge, incomplete verification is studied, aiming to solve the relaxed problem, i.e., computing the lower bound of $\cM$ which is more tractable.  However, the relaxations in existing approaches are typically too loose, which induces another problem identified as the \textbf{tightness challenge}. 
        For example, the widely-used linear relaxations are shown significantly looser than complete verification in practice~\cite{salman2019convex}.
        Theoretically, if complete verification can certify robustness radius $\epsilon_0$, unless $\NP=\P$, there is no polynomial-time verification that can guarantee a constant fraction between its certified robustness radius and $\epsilon_0$~\cite{weng2018towards}.
        The trade-off between scalability and tightness, i.e., either scalability or tightness can be achieved but not both, constitutes the main obstacle for robustness verification.


        \para{Robust training.}
        Given the scalability and tightness challenges, vanilla DNNs are challenging to verify, where verification approaches either need a long running time or output trivial bounds.
        To enhance the certifiability, many robust training approaches are proposed, which are typically related to or derived from corresponding verification by optimizing verification-inspired regularization terms or injecting specific data augmentation during training. 
        In practice, after robust training, the model usually achieves high certified robustness.
        Thus, robust training is a strong complement to robustness verification approaches.
        
        \begin{figure}[!t]
            \centering
            \includegraphics[width=0.95\linewidth]{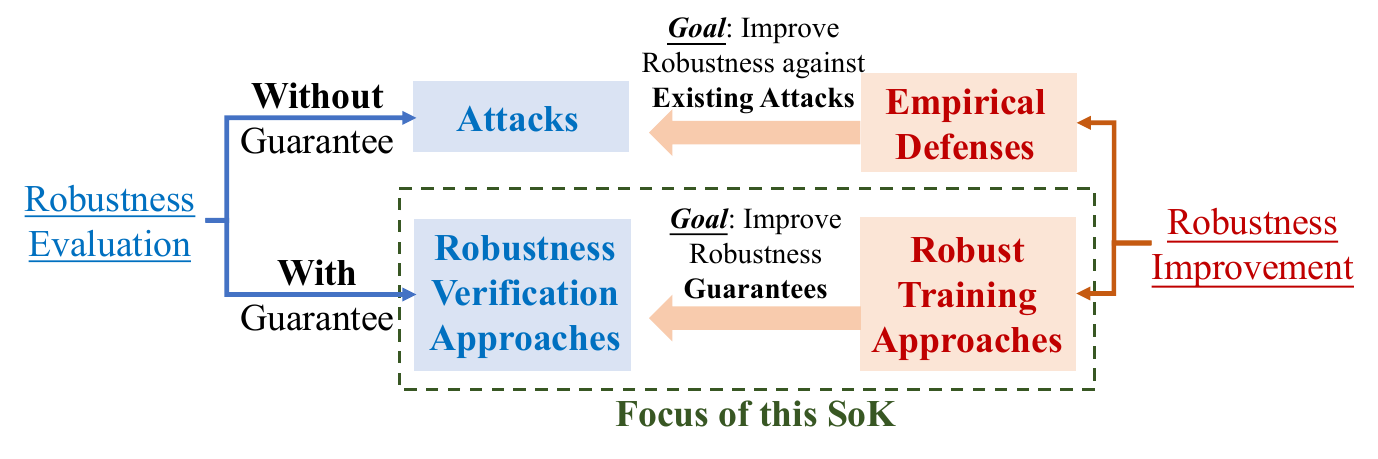}
            \vspace{-1.3em}
            \caption{Different approaches for evaluating and improving DNN robustness against evasion attacks.}
            \vspace{-2.0em}
            \label{fig:categorization-high-level}
        \end{figure}

    \begin{figure*}
        \centering
        \includegraphics[width=0.95\textwidth]{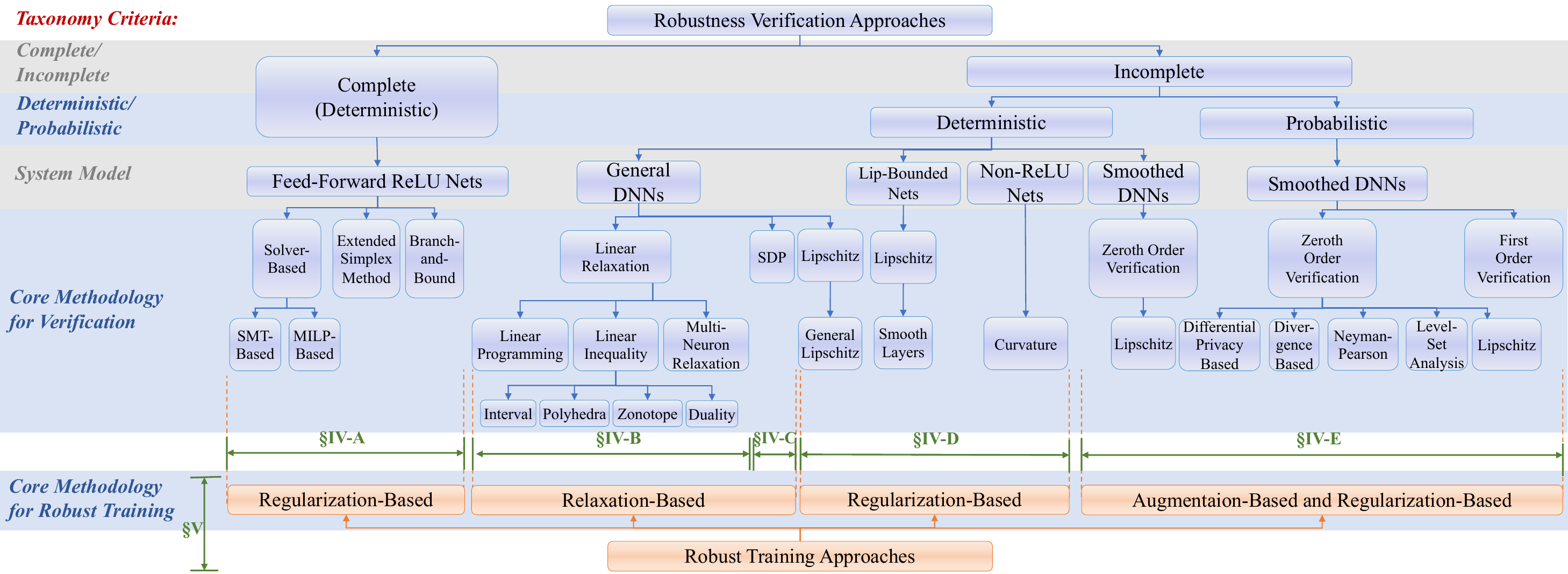}
        \vspace{-1em}
        
        \caption{\textbf{Taxonomy} of certifiably robust approaches.
        \textcolor{darkblue}{Blue boxes} show the taxonomy of verification approaches.
        \textcolor{orange}{Orange boxes} show the taxonomy of robust training approaches, and vertical dotted lines show the suitable verification for corresponding training approaches.
        Left columns list taxonomy criteria per level.
        Sections illustrating corresponding approach category are shown as \textcolor{darkgreen}{green segments}.
        }
        \label{fig:tree-categorization}
        \vspace{-0.8em}
    \end{figure*}
    \input{tables/property-table}

        \para{Relationship with empirical attacks and defenses.}
        Towards evaluating and improving DNN robustness, another active line of research is attacks and empirical defenses.
        Strong white-box attacks, such as CW attack~\cite{carlini2017towards}, PGD attack~\cite{madry2017towards}, and AutoAttack~\cite{croce2020reliable}, are widely used to evaluate DNN robustness~(e.g., \cite{yangli2021trs,zhang2019theoretically}).
        To improve model robustness against these attacks, many empirical defenses are proposed, such as adversarial training~\cite{gowal2021improving,madry2017towards,pang2020bag,Wong2020Fast} and TRADES~\cite{zhang2019theoretically}.
        As illustrated in \Cref{fig:categorization-high-level}, both attacks and verification approaches can be used to evaluate DNN robustness, but verification approaches can provide robustness guarantees against any possible future attacks;
        both empirical defenses and robust training approaches can improve DNN robustness, but empirical defenses aim to improve robustness against existing attacks and robust training approaches aim to improve robustness guarantees.
        We note that:
        (1)~The strongest attack~(which always discovers adversarial example if exists) is the strongest verification~(complete verification).
        For robustness evaluation, attack and verification can be viewed as approaching from two sides~(over-estimation and under-estimation) to the same goal~(precise evaluation).
        (2)~Complete verification approaches can be used to evaluate and compare empirical defenses on small models~(not on large models due to scalability challenges). 
        In practice, models trained with strong empirical defenses can be certified to have high robustness by complete verification~\cite{wang2021beta}.
        In contrast, most incomplete verification cannot certify high robustness for empirically defended models.
        More discussion is in \Cref{sec:training-approach}.

\section{Taxonomy of Certifiably Robust Approaches}
    \vspace{-0.5em}
    \label{sec:main-taxonomy}

    In this section, we provide a comprehensive taxonomy of existing robustness verification and robust training approaches~(\Cref{fig:tree-categorization}), and characterize their properties~(\Cref{tab:property-table}).

    \para{Taxonomy of robustness verification and robust training.}
    In \Cref{fig:tree-categorization}, we present a taxonomy of existing robustness verification and robust training approaches.
    In the taxonomy, the \underline{first-level} is ``complete vs. incomplete'', and the \underline{second-level} is ``deterministic vs. probabilistic''. These concepts are as defined in \Cref{subsec:target-model}.
    Note that there is no complete and probabilistic verification approach yet.
    In the \underline{third-level}, we categorize verification approaches based on the system model.
    Under the third level, we categorize verification approaches by their core methodologies.
    We will illustrate verification approaches in detail in \Cref{sec:verification-approaches}.
    %
    %
    The robust training approaches are shown in orange.
    Based on their core methodologies, there are three categories: regularization-based, relaxation-based, and augmentation-based approaches.
    We will illustrate robust training approaches in detail in \Cref{sec:training-approach}.
    
    \para{Properties of verification approaches.}
    In \Cref{tab:property-table}, we summarize the key properties of each verification approach, including the system model, the supported $\ell_p$ adversary types, scalability, and its verification tightness.
    
    For the \textbf{supported $\ell_p$} in \Cref{tab:property-table},  ``$\checkmark$'' means well-supported $\ell_p$ adversaries, ``$(\checkmark)$'' means supported adversaries but the verification is not as tight as others, and empty means unsupported adversaries.
    To measure the \textbf{scalability}, we use ``\textit{the largest dataset~(in terms of input dimension) that has been demonstrated feasible to certify by existing work using the corresponding verification approach under radius $\epsilon \ge 1/255$}'' as the criterion. The threshold $1/255$ is the smallest considered $\epsilon$ we have seen in the literature for these image datasets.
    The dataset effectively measures scalability.
    For example, the approach scaling up to ImageNet is more scalable than the one to MNIST.
    We also provide a quantitative measure: the best known time complexity for verifying an arbitrary input, given an arbitrary network with depth $l$, width $w$ in terms of neurons, and sampling number $S$~(for smoothed DNNs).
    For \textbf{tightness}, the tightest verification approaches are complete ones.
    For incomplete approaches, we rank the tightness by our benchmark results shown in \Cref{sec:evaluation}, or empirical observations and theoretical results from published papers.
    General DNNs are ranked by $T_n$ and smoothed DNNs are ranked by $ST_n$ where larger $n$ means tighter approaches. $T_n$- and $ST_n$-denoted approaches are incomparable since their system models are different. 
    More discussion on the scalability and tightness measurements are in App.~\ref{adxsec:scalability-tightness}.
    
    From \Cref{tab:property-table}, we observe that for general DNNs, complete or tight deterministic approaches~($\ge T_5$) can only handle CIFAR-10-sized models, and only looser verification can go beyond this scale.
    For large ImageNet-sized models, verification for general DNNs cannot support such a scale yet; only approaches for smoothed DNNs can, while they cannot provide nontrivial verification against $\ell_\infty$ adversary yet.
    This reflects the fundamental trade-off between scalability and tightness challenge.
    We will discuss this further in \Cref{sec:evaluation}.

    \vspace{-0.3em}
\section{Robustness Verification Approaches}
    \vspace{-0.3em}
    \label{sec:verification-approaches}
    We illustrate representative verification approaches in this section: complete verification~(\Cref{sec:complete-verification}); incomplete verification, including linear relaxation-based~(\Cref{sec:lp-relaxation-intro}), SDP~(\Cref{sec:sdp-intro}), Lipschitz-/curvature-based~(\Cref{sec:lipschitz-intro}),
    and probabilistic approaches~(\Cref{sec:probabilistic-robustness-verification-root}).
    We conclude each subsection by highlighting the implications.
    We summarize practical and research implications in \Cref{subsec:guideline,sec:discussion} respectively.

    \vspace{-0.3em}
 \subsection{Complete Verification}
    \label{sec:complete-verification}
    \vspace{-0.3em}
 
    Here we illustrate complete robustness verification approaches, which usually consider $\cL_\infty$ adversary and support feed-forward ReLU networks~(see \Cref{def:feed-forward-nn}).
    All these complete verification approaches have worst-case exponential time complexity due to the hardness of verification~\cite{katz2017reluplex,weng2018towards}, but some of them perform well in practice, being able to verify DNNs with several thousands of neurons~\cite{tjeng2018evaluating,wang2021beta}.
    Many complete verification approaches rely on the neuron activation patterns, so below we first categorize neurons by their activation patterns.
    \begin{definition}[Stable and Unstable Neurons]
        Let $z=\relu(\hat z)$ be a neuron in a feed-forward ReLU network.
        For a given input $x$, if the input $\hat z < 0$, we call $z$  \emph{inactive}, otherwise \emph{active}. 
        Let $\cS$ be an input region, for any $x \in \cS$, if we can certify that input $\hat z$ is always $> 0$ or $\le 0$, we call neuron $z$ \emph{stable}
        in region $\cS$; otherwise we call $z$ \emph{unstable}.
        \label{def:active-inactive-relu-neuron}
    \end{definition}
    \begin{remark*}
        When a neuron $z$ is stable, it serves as a linear mapping $x\mapsto 0$~(inactive neuron) or $x\mapsto x$~(active neuron).
    \end{remark*}
    
    \subsubsection{\textbf{Solver-Based Verification}\texorpdfstring{~\cite{cheng2017maximum,dutta2017output,lomuscio2017approach,tjeng2018evaluating,pulina2010abstraction,pulina2012challenging}}{}}
        \label{subsec:solver-based}
        By inspecting the definition of feed-forward neural networks~(\Cref{def:feed-forward-nn}), we can observe that the DNN $f_\theta(x)$ is defined by the sequential composition of affine transformations and ReLU operations.
        Both affine transformations and ReLU operations can be encoded by a conjunction of linear inequalities.
        For example, 
        $z_{i,j} = \relu(\hat z_{i,j}) \iff ((\hat z_{i,j} < 0) \wedge (z_{i,j} = 0)) \vee ((\hat z_{i,j} \ge 0) \wedge (z_{i,j} = \hat z_{i,j}))$.
        Thus, general-purpose SMT solvers such as Z3~\cite{de2008z3} can be directly applied to solve the satisfiability problem of boolean predicate $\bigwedge_{y' \in [C]: y' \neq y_0} (\gM(y_0,y') > 0)$~(see \Cref{prob:verification-optimization-problem}), which yields a solution to complete verification.
        However, SMT-based verification is generally not scalable~\cite{pulina2010abstraction,pulina2012challenging} and can verify DNNs with only hundreds of neurons, which is too small even for the simple MNIST dataset. 
        
        Another way is to encode the verification problem as a mixed-integer linear programming~(MILP) problem.
        In MILP, the constraints are linear inequalities and the objective is a linear function.
        However, different from linear programming~(LP), in MILP we can constrain some variables to take only integer values instead of real numbers.
        This additional expressive power allows MILP constraints to encode the non-linear ReLU operations and the whole DNN model~\cite{cheng2017maximum,lomuscio2017approach}.
        Thus, the verification problem can be precisely encoded as an MILP problem.
        By leveraging efficient MILP solvers such as \t{Gurobi}~\cite{gurobi}, MILP-based verification is feasible on medium-sized CIFAR-10 models if the model is specifically trained to favor certifiability~\cite{wong2018scaling,gowal2019scalable,zhang2020towards,konig2021speeding}.
        However, the naturally trained or empirically defended DNNs are still hard to verify by these approaches even on MNIST~\cite{tjeng2018evaluating}.
    
    \subsubsection{\textbf{Extended Simplex Method}\texorpdfstring{~\cite{katz2017reluplex,katz2019marabou}}{}}
        The DNN model is composed of affine transformations and ReLU operations which correspond to linear constraints and ReLU constraints respectively.
        When there are only linear constraints, the verification problem is a linear programming problem and can be effectively solved by the simplex method~\cite{ignizio1994linear}.
        In \cite{katz2017reluplex,katz2019marabou}, the simplex method is extended to handle ReLU constraints.
        The core idea is to iteratively check whether the ReLU constraints are violated and fix them.
        If the violation cannot be easily fixed, we split the neuron into active and inactive and solve subproblems respectively.
        
    \subsubsection{\textbf{Branch-and-Bound}\texorpdfstring{~\cite{bak2020improved,bunel2019branch,bunel2018unified,ehlers2017formal,ferrari2021complete,fromherz2021fast,gehr2018ai2,jordan2019provable,palma2021scaling,wang2018formal,wang2018efficient,wang2021beta,zhang2022general}}{}}
        \label{subsec:branch-and-bound}
        Another line of complete verification is branch-and-bound.
        Most competitors in VNN-COMP, an annual DNN verification competition, build their verification tools based on branch-and-bound~\cite{bak2021second}, and
        the winner tool of VNN-COMP 2021 and 2022, \t{$\alpha$-$\beta$-CROWN}~\cite{wang2021beta,zhang2022general}, is based on branch-and-bound.
        The branch-and-bound verification relies on the \textbf{piecewise-linear property} of DNNs:
        %
        %
            %
        %
        Since each ReLU neuron outputs $\relu(x) = \max\{x,\,0\}$, it is always locally linear within some region around input $x$.
        Since feed-forward ReLU networks are the composition of these piecewise linear neurons and (linear) affine mappings, \emph{the output is locally linear w.r.t. input $x$.}
        This property is formally stated and proved in \cite{jordan2019provable}.
        It serves as the foundation for branch-and-bound verification.

        Given an input $x_0$ with true class label $y_0$ and perturbation radius $\epsilon$, recall that the verification problem can be reduced to deciding whether $\gM(y_0,y') > 0$ for any $y' \in [C] \setminus \{y_0\}$~(see \Cref{prob:verification-optimization-problem}).
        A branch-and-bound verification approach first applies incomplete verification to derive a lower bound and an upper bound of $\gM(y_0,y')$: if the lower bound is positive then terminate with ``verified''; if the upper bound is non-positive then terminate with ``not verified''---\textbf{bounding}. Otherwise, the approach recursively chooses a neuron $z_{i,j} = \relu(\hat z_{i,j})$ to split into two branches: $\hat z_{i,j} < 0$~(inactive branch) and $\hat z_{i,j} \ge 0$~(active branch)---\textbf{branching}.
        For inactive branch, we have the constraint $\hat z_{i,j} < 0, z_{i,j} = 0$;
        and for active branch, we have the constraint $\hat z_{i,j} \ge 0, z_{i,j} = \hat z_{i,j}$.
        Therefore, for each branch the neuron brings only linear constraints, and we again apply incomplete verification to determine whether $\gM(y_0,y') > 0$ for each branch.
        If for both branches, we can verify that $\gM(y_0,y')$ is always positive/negative or the branching condition is infeasible, the verification terminates; otherwise, we further split other neurons recursively.
        When all neurons are split, the branch will contain only linear constraints, and thus the approach applies linear programming to compute the precise $\gM(y_0,y')$ and verify the branch.
        The branch-and-bound framework is formalized in \cite{bunel2019branch,bunel2018unified,ehlers2017formal}, and opens a wide range of design choices, leading to approaches with different implementation and scalabilities.
        Some verification approaches efficiently traverse the piecewise linear regions around the clean input $x_0$ to exhaustively search adversarial examples in the region $B_{p,\epsilon}(x_0)$~\cite{bak2020improved,fromherz2021fast,jordan2019provable}, which work better under $\ell_2$ adversary while other branch-and-bound approaches work better under $\ell_\infty$ adversary, 
        because under $\ell_2$ adversary input region has special geometric properties that can be exploited for traversal-based approaches~\cite{fromherz2021fast}.

        \begin{practitioner}
            Though complete verification approaches have worst-case exponential time complexity, they, especially branch-and-bound approaches such as \t{$\alpha$-$\beta$-CROWN}~\cite{wang2021beta,zhang2022general}, can verify DNNs with up to $10^5$ neurons like ResNet~\cite{he2016deep} in practice within several minutes, if the model is specifically trained to favor certifiability.
            This model size is of moderate level on CIFAR-10.
            For models that are not specifically trained, complete verification can handle DNNs with up to $10^4$ neurons and roughly $6$ layers, corresponding to small models on CIFAR-10 and large models on MNIST.
            Therefore, for simple tasks and simple DNNs, such as those for aircraft control systems~\cite{bak2020improved,julian2016policy,katz2017reluplex,katz2019marabou,singh2018robustness,shriver2021dnnv,wang2018efficient}, it is feasible to deploy complete verification to verify the robustness.
            The branch-and-bound approaches are more scalable than solver-based and extended simplex approaches.
            A more comprehensive comparison of existing verifiers can be found in VNN-COMP~2021~\cite{bak2021second} and 2022~\cite{muller2022third} report.
        \end{practitioner}
        
        \begin{researcher}
            For complete verification, the primary research goal is to develop more scalable approaches.
            For solver-based verification, it is important to find a more solver-friendly problem encoding or design specific optimizations tailored for DNN verification inside the solver~\cite{konig2021speeding}.
            For branch-and-bound verification, it is a popular direction to find an efficient incomplete verification heuristic for the bound computation that has a better trade-off between tightness and efficiency~\cite{bunel2019branch,palma2021scaling,wang2021beta,zhang2022general}.
            In addition, finding a new branching heuristic, either rule-based or learning-based, is a promising direction to boost the scalability of complete verification.
            
            Recently, Zombori~et~al~\cite{zombori2020fooling} and Jia and Rinard~\cite{jia2021exploiting} discovered that some complete verification approaches are unsound under floating-point arithmetic and such unsoundness can be exploited to fool the verifier.
            Thus, future approach developers should consider floating-point rounding errors.
        \end{researcher}
 
 \subsection{Incomplete Verification via Linear Relaxation}
 
    \label{sec:lp-relaxation-intro}
    Due to the scalability barrier of complete verification, many incomplete verification approaches based on relaxations are proposed.
    Among them, linear relaxations are well studied.
    This category of verification approaches runs much faster and many can scale up to large ResNet models on Tiny ImageNet, which contain around $10^5$ neurons~\cite{xu2020automatic}.
    
    Linear relaxation based approaches rely on ReLU polytope, which we define below and illustrated in App.~\ref{adxsubsec:single-illustration}.
    
        \begin{definition}[Polytope for Unstable ReLU]
            \label{def:tightest-linear-polytope}
            For neuron $z_{i,j} = \relu(\hz_{i,j})$, let $l_{i,j}$ and $u_{i,j}$ be the lower bound and upper bound of its output when the input region is $B_{p,\epsilon}(x_0)$:
            \begin{equation}
                \small
                l_{i,j} \le \min_{x\in B_{p,\epsilon}(x_0)} \hz_{i,j}(x) \le
                \max_{x\in B_{p,\epsilon}(x_0)} \hz_{i,j}(x) \le u_{i,j}.
                \label{eq:l-u-bound-for-relu}
            \end{equation}
            Then, if $l_{i,j} < 0 < u_{i,j}$, the \textbf{unstable neuron} $z_{i,j} = \relu(\hz_{i,j})$ can be bounded by following linear constraints:
            \begin{equation}
                \small
                z_{i,j} \ge 0,\, z_{i,j} \ge \hz_{i,j},\, z_{i,j} \le \dfrac{u_{i,j}}{u_{i,j} - l_{i,j}} (\hz_{i,j} - l_{i,j}).
                \label{eq:linear-relu-0}
            \end{equation}
            These constraints define a region called \emph{ReLU polytope}.
        \end{definition}
    %
    %
        When both $l_{i,j}$ and $u_{i,j}$ are tight, 
        the polytope is the tightest convex hull for this neuron.
        For stable ReLU, linear constraint $z_{i,j}=0$ or $z_{i,j} = \hat z_{i,j}$ defines its linear relaxation.

    In general, all linear relaxation based approaches require computing $l_{i,j}$ and $u_{i,j}$~(see \Cref{def:tightest-linear-polytope}) for each neuron $z_{i,j}$, then they compute an over-approximation bound $\cS$ for the region $f_\theta \left(B_{p,\epsilon}(x_0)\right) := \{ f_\theta(x):\, x\in B_{p,\epsilon}(x_0) \}$, i.e., $\cS \supseteq f_\theta \left( B_{p,\epsilon}(x_0) \right)$.
    $\cS$ is described by linear constraints so that it is easy to verify whether all points in $\cS$ lead to the true class $y_0$.
    If it is true, the region $f_\theta \left(B_{p,\epsilon}(x_0)\right)$ is robust. 
    %
    
 
    \subsubsection{\textbf{Linear Programming}\texorpdfstring{~\cite{weng2018towards,salman2019convex}}{}}
    \label{sec:linear-programming-intro}
        Based on \Cref{def:tightest-linear-polytope}, we can directly use the polytope shown in \Cref{fig:tightest-linear-polytope} in App.~\ref{adxsubsec:single-illustration} as the relaxation for verification, which results in the approach named \t{LP-full}~\cite{weng2018towards,salman2019convex}.
%
%
        In \t{LP-full},  $l$ and $u$ are computed layer by layer,  the polytope relaxation (\Cref{def:tightest-linear-polytope}) is then applied for each ReLU neuron, and finally, the verification is performed by solving the resulting linear programming~(LP) problem. 
        Due to the relaxation, we obtain a lower bound of $\cM(y_0,y')$ in \Cref{prob:verification-optimization-problem}. 
        Even though LPs can be solved in polynomial time, in practice, solving LP is still expensive.
        Applying \t{LP-full} on a typical model on CIFAR-10 for verifying a single instance takes several hours to several days~\cite{salman2019convex}.
        Moreover, although \t{LP-full} is the tightest verification using single neuron linear relaxations, compared with complete verification, the certified robustness radius $\epsilon$ is usually $1.5 - 5$ times smaller, which indicates the intrinsic tightness barrier of linear relaxations.
    \subsubsection{\textbf{Linear Inequality}\texorpdfstring{~\cite{anderson2019optimization,dvijotham2018training,dvijotham2018dual,gowal2019scalable,lyu2020fastened,lyu2021towards,mirman2018differentiable,singh2019abstract,singh2018robustness,weng2018towards,wong2018provable,wong2018scaling,xu2020automatic,zhang2018efficient}}{}}
        \label{sec:linear-inequality-propagation-intro}
        To circumvent solving expensive LP, further relaxations are applied, which can be divided into \emph{interval} bound propagation~(IBP), \emph{polyhedra} abstraction, \emph{zonotope} abstraction, and \emph{duality}-based approaches.
        
        \para{Inteval bound propagation~(IBP).}
        \label{sec:ibp-intro}
        A more straightforward and efficient but much looser approach comes from directly propagating $l$ and $u$ defined in \Cref{eq:l-u-bound-for-relu} through the layers of the given DNN model.
        Given perturbed input region $B_{p,\epsilon}(x_0)$, for the first layer, we have $z_1 = x \in [x_0 - \epsilon, x_0 + \epsilon]$.
        We let $[l_1, u_1]$ to represent this numerical interval for the first layer $z_1$.
        Then, we derive $[l_{i+1},\,u_{i+1}]$ for layer $z_{i+1}$ from $[l_i,\,u_i]$:
        If $l_k \le z_k \le u_k$, based on $\hz_{k+1} = \bW_k z_k + b_k$, $\hz_{k+1}$ can be bounded by $\hat l_{k+1} \le \hz_{k+1} \le \hat u_{k+1}$ where
        $$
            \small
            \hat l_{k+1} = \bW_k^+ l_k + \bW_k^- u_k + b_k,
            \, 
            \hat u_{k+1} = \bW_k^+ u_k + \bW_k^- l_k + b_k.
        $$
        $\bW^+$ sets negative elements in $\bW$ to $0$; and $\bW^-$ sets positive elements to $0$.
        Then, $z_{k+1} = \relu(\hz_{k+1})$ can be bounded by $l_{k+1} = \max\{\hat l_{k+1},\,0\} \le z_{k+1} \le \max\{\hat u_{k+1},\,0\} = u_{k+1}$.
        
        Through each layer, this bound propagation performs only four matrix-vector products, which are in the same order of model inference.
        As a result, the approach is very scalable for verifying large models on ImageNet, but on ImageNet it yields trivial bounds due to its looseness.
        The approach 
        is called IBP~\cite{gowal2019scalable} or interval arithmetic~\cite{wang2018formal}.
        
        As we will discuss in \Cref{sec:training-approach}, though for normal DNNs, IBP is usually loose.
        For the models that are specifically trained with IBP, IBP can verify close-to-best certified robustness among linear-relaxation-based approaches.
        Some work~\cite{jovanovic2021certified} conjectures that IBP bound, though loose, is smoother than other linear relaxations and thus more suitable for training.
        
        \para{Polyhedra abstraction.}
        The polyhedra abstraction based verification approaches, such as \t{Fast-Lin}~\cite{weng2018towards}, \t{CROWN}~\cite{zhang2018efficient}, and \t{DeepPoly}~\cite{singh2019abstract},
        replace the two lower bounds in the ReLU polytope shown in \Cref{eq:linear-relu-0} by a single lower bound, resulting in one lower and one upper bound for each neuron respectively.
        The idea is illustrated in \Cref{fig:relu-lower-bound-1,fig:relu-lower-bound-2,fig:relu-lower-bound-3} in App.~\ref{adxsubsec:single-illustration}.
        The advantage of using a single linear lower bound is that:
        (1)~the linear bounds can be propagated through  layers efficiently instead of solving LP problem---the verification is more scalable than LP;
        and (2)~linear bounds maintain interactions between different components to some degree---the verification is typically tighter than IBP.
        We call these approaches ``polyhedra abstraction based'' approaches since they essentially compute polyhedra domain abstraction interpretation~\cite{gehr2018ai2} for DNNs.
        We defer technical details along with the illustration of zonotope abstraction and duality-based approaches to App.~\ref{adxsec:polyhedra-illustration}.
        
        For all linear inequality based verification approaches,
        Salman~et~al~\cite{salman2019convex} prove the \emph{convex barrier}: these approaches cannot be tighter than linear programming based approaches~(introduced in \Cref{sec:linear-programming-intro}).

    \subsubsection{\textbf{Multi-Neuron Relaxation}\texorpdfstring{~\cite{muller2022prima,palma2021scaling,singh2019beyond,tjandraatmadja2020convex}}{}}
        To circumvent the convex barrier mentioned above, Singh~et~al~\cite{singh2019beyond} and Tjandraatmadja~et~al~\cite{tjandraatmadja2020convex} found that for ReLU that takes multiple input variables~(e.g., $z = \relu(x+y)$ takes scalars $x$ and $y$), if considering multiple input variables together, the tightest convex polytope is tighter than applying single-neuron polytope~(\Cref{def:tightest-linear-polytope}) along the base direction~(i.e., $(x+y)$-direction for $z = \relu(x+y)$).
        \Cref{fig:multi-neuron} in App.~\ref{adxsec:multi-neuron-illustration} illustrates this observation.
        Multi-neuron relaxation based approaches are proposed to leverage the multivariate convex relaxations to tighten the verification.
        Among them, \t{k-ReLU}~\cite{singh2019beyond} and \t{PRIMA}~\cite{muller2022prima} consider $k$~($k\le 5$) inputs at once.
        Tjandraatmadja~et~al~\cite{tjandraatmadja2020convex} point out that tightest convex polytope may contain exponential number of linear constraints, 
        and propose \t{C2V} to heuristically find out and only preserve more useful constraints.
        \t{Active-Set}~\cite{palma2021scaling} improves upon \t{C2V} with gradient-based optimization and better heuristics on constraint selection.
        \t{GCP-CROWN}~\cite{zhang2022general} extracts convex constraints from MILP solvers and integrate them in linear inequality propagation, which can be viewed as leveraging multi-neuron relaxations in branch-and-bound complete verification.

        \begin{practitioner}
            When the DNN model is too large to be verified by complete verification approaches, linear relaxation based approaches are good options.
            Among these approaches, IBP is the most scalable and typically the loosest one~\cite{gowal2019scalable}, yielding trivial bounds on normal DNNs.
            However, on models that are specifically trained for IBP, the IBP certified robustness can be quite satisfactory on large CIFAR-10 and Tiny ImageNet datasets~\cite{xu2020automatic,shi2021fast,zhang2020towards}.
            Multi-neuron relaxation based approaches are the tightest but least scalable ones.
            Effective heuristics enable multi-neuron relaxation based approaches to significantly improve the scalability at small tightness loss, so they can verify decent robustness on medium-sized CIFAR-10 models with around $10^5$ neurons more efficiently than complete verification~\cite{muller2022prima}.
            But for very deep neural networks~(layers $\ge 10$), due to the amplification of over-approximation, linear relaxation based approaches cannot certify nontrivial robustness.
            
            Besides those mentioned above, there are linear relaxation based verification approaches and implementations aiming to efficiently support specific DNN architectures, including CNN~\cite{boopathy2019cnn}, residual blocks~\cite{wong2018scaling}, and other activation functions~\cite{muller2022prima,xu2020automatic,zhang2018efficient}.
        
            
            
        \end{practitioner}
        
        \begin{researcher}
            For linear relaxation based approaches, a tighter and more scalable multi-neuron relaxation based approach is in need.
            Due to the worst-case exponential number of linear constraints in the tightest convex relaxation, it is important to develop efficient heuristics to synthesize the critical constraints and solve the verification problem with these constraints.
            Both rule-based heuristics performance~\cite{muller2022prima,tjandraatmadja2020convex,zhang2018efficient} and learning-based heuristics~\cite{paulsen2022linsyn} are shown effective and are promising directions.
            Extracting constraints from existing solvers is effective as well~\cite{zhang2022general}.
            Furthermore, towards the ultimate goal of improving certified robustness, it is also an important topic to develop more efficient robust training approaches for linear relaxation based verification, which we will discuss more~in~\Cref{sec:training-approach}.
        \end{researcher}
    
 \subsection{Incomplete Verification via SDP}
    \label{sec:sdp-intro}
    
    
    Semidefinite programming (SDP) can be applied for incomplete verification: \t{Verify}~\cite{raghunathan2018semidefinite} formulates the robustness verification as an SDP problem, which is a convex optimization problem, where the decision variable is a symmetric and semi-positive matrix whose elements can be linearly constrained.
    The key formulation in \t{Verify} is
    \vspace{-0.7em}
    \begin{equation}
        \small
        z_{ij} = \relu(\hz_{ij}) \Longleftrightarrow
        \left\{
        \begin{aligned}
            & z_{ij} \ge 0, \quad z_{ij} \ge \hz_{ij}, \\
            & z_{ij}z_{ij}^\T = z_{ij} \hz_{ij}^\T. \\
        \end{aligned}
        \right.
        \label{eq:relu-sdp-formulation}
        \vspace{-0.7em}
    \end{equation}
    To handle the quadratic constraint, it defines the vector $v := (1,\,z_1,\,\dots,\,z_l)^\T$ which encodes all ReLU activations, and considers the matrix $\bP = vv^\T$ as the SDP decision variable. 
    Since we replace the constraint $\rank(\bP)=1$ coming from $\bP = vv^\T$ by semidefinite constraint $\bP \succeq 0$, it is a relaxation.
    Then, the constraints in \Cref{eq:relu-sdp-formulation} can be directly treated as linear inequality constraints on $\mP$'s elements and thus can be precisely encoded by the SDP problem along with the optimization objective for verification~(\Cref{prob:verification-optimization-problem}).
    Several variants of SDP encoding are studied~\cite{raghunathan2018certified,dvijotham2019efficient,fazlyab2019safety}.
    
    
    \begin{practitioner}
        SDP-based verification approaches are incomplete but quite tight---empirically the tightness lies between linear programming based and multi-neuron relaxation based approaches.
        However, the utility of SDP-based verification approaches is limited by the slow SDP solving process.
        Even with the specialized solver as in \cite{dathathri2020enabling}, verifying a thousand-neuron level DNN model requires several hours to one day, while complete verification and linear relaxation based approaches can terminate within minutes.
        Therefore, 
        complete verification or linear relaxation based approaches are recommended instead of SDP-based verification at the current stage.
    \end{practitioner}
    
    \begin{researcher}
        The major challenge of SDP-based verification is its scalability.
        The improvements for SDP-based verification would come from either more efficient and tighter relaxation~\cite{batten2021efficient} or more efficient solvers~\cite{dathathri2020enabling}.
        For example, the specialized first-order solver proposed in \cite{dathathri2020enabling} greatly boosts the scalability.
        Further scalability improvements need to be explored.
    \end{researcher}
    
 \subsection{Incomplete Verification via Lipschitz or Curvature Bounds}
    \label{sec:lipschitz-intro}
    Some verification approaches use the Lipschitz bound or curvature bound of DNN function $f_\theta$ to verify its robustness.
    
    \begin{definition}[Lipschitz Constant]
        \label{def:lipschitz-constant}
        We say scalar function $g: \bbR^n \supseteq \cX \to \sR$ has local \emph{Lipschitz constant} $L$ w.r.t. $\cL_q$ norm in region $B_{p,\epsilon}(x_0)$ if $\forall x_1, x_2 \in B_{p,\epsilon}(x_0)$,
            $\frac{|g(x_1) - g(x_2)|} {\|x_1 - x_2\|_q} \le L$.
    \end{definition}
    
    \noindent
    We can lower bound $f_\theta(x)_{y_0} - f_\theta(x)_{y'}$ for any $x\in B_{p,\epsilon}(x_0)$ given Lipschitz constant, and thus certify robustness~\cite{szegedy2013intriguing}.

    

    \subsubsection{\textbf{General Lipschitz}\texorpdfstring{~\cite{lee2020lipschitz,globally2021leino,singla2021fantastic,szegedy2013intriguing,tsuzuku2018lipschitz,weng2018towards,zhang2019recurjac}}{}}
        \label{sec:general-lipschitz}
        Some verification approaches aim at computing a tight Lipschitz bound for general neural networks, and we call them \emph{general Lipschitz based verification approaches}.
        A commonly-used approach~\cite{szegedy2013intriguing,tsuzuku2018lipschitz,globally2021leino} is to compute a global Lipschitz constant w.r.t. $\ell_2$ norm by multiplying the spectral norm of all weight matrices $L = \prod_{i=1}^{l-1} \|\mW_i\|$ of the ReLU neural network, where the spectral norm can be computed by power iteration algorithm~\cite{mises1929praktische}.
        Efforts have been made on tightening this bound~\cite{singla2021fantastic}.
        The global Lipschitz constant is usually too loose to provide nontrivial certified robustness in practice, but it can be efficiently regularized during training.
        When this constant is regularized, this verification can bring non-trivial certified robustness against $\ell_2$ norm, which we will discuss in detail in \Cref{sec:training-approach}.
        Global Lipschitz constant computation is efficient and thus scalable to models on Tiny ImageNet dataset~\cite{lee2020lipschitz,globally2021leino}.
        
        Global Lipschitz bound can be improved by finer-grained analysis on convolutional layers~\cite{lee2020lipschitz}, by computing local Lipschitz bound~\cite{fazlyab2019efficient,weng2018towards,zhang2019recurjac},
        or by combining with IBP~\cite{lee2020lipschitz}.
        Currently, general Lipschitz based verification can certify nontrivial robustness against only $\ell_2$ adversary.
        
    \subsubsection{\textbf{Smooth Layers}\texorpdfstring{~\cite{hein2017formal,li2019preventing,singla2021householder,trockman2021orthogonalizing,zhang2021towards}}{}}
        \label{sec:smooth-layers}
        Besides general Lipschitz based verification, another thread of research proposes specific layer structures which we call \emph{smooth layers} and proves Lipschitz constant for these layer structures.
        For example, there are different designs of orthogonal convolutional layers~\cite{li2019preventing,trockman2021orthogonalizing}.
        They usually use parameterization or transformation to explicitly construct trainable convolutional layers which are orthogonal and thus have $1$ as the Lipschitz constant.
        When this small Lipschitz constant is proved, general Lipschitz based verification can provide robustness certification.
        However, these approaches are restricted to $\ell_2$ adversary.
        Recently, Zhang~et~al~\cite{zhang2021towards} propose a novel activation function $x \mapsto \|x-w\|_\infty + b$ which is called $\ell_\infty$ neuron and is $1$-Lipschitz w.r.t. $\ell_\infty$ norm.
        This design enables general Lipschitz based verification to certify robustness against $\ell_\infty$ adversary.
        Combined with effective training~\cite{zhang2021boosting}, this approach can certify state-of-the-art $\ell_\infty$ certified robustness.
        
    \subsubsection{\textbf{Curvature}\texorpdfstring{~\cite{singla2020second}}{}}
    \label{sec:curvature-intro}
    If a DNN uses activation functions that have non-zero second-order derivatives, such as sigmoid, Singla and Feizi~\cite{singla2020second} propose an efficient algorithm to bound the DNN's curvature, i.e., second-order derivatives.
    Based on the curvature bound, we can compute a lower bound of \Cref{prob:verification-optimization-problem} and thus certify the model's robustness.
    Compared with others, curvature-based verification cannot be applied to the widely-used ReLU networks and can only certify against $\ell_2$ adversary, so the application scenario is a bit limited. 
    Though on small dataset like MNIST, this approach can verify high robustness for some robustly trained models.
        \begin{practitioner}
            When applying Lipschitz- or curvature-based approaches to general DNN models, the resulting robustness verification is usually trivial.
            However, if the model is regularized to have a small Lipschitz constant or uses smooth layers with effective training, these approaches can provide nontrivial bounds on both small and large datasets~(MNIST, CIFAR-10, and Tiny ImageNet) against both $\ell_2$ and $\ell_\infty$ adversaries.
        \end{practitioner}
        
        \begin{researcher}
            Among these Lipschitz and curvature based verification,
            we believe that the potential of smooth layers based verification has not been fully explored yet.
            For example, many orthogonal training methods can lead to smooth layers~(e.g., \cite{bansal2018can,huang2020controllable,wang2020orthogonal}), and smooth variants may exist for recent architectures such as transformers~\cite{vaswani2017attention}.
            Improving on these directions may lead to state-of-the-art certified robustness.
        \end{researcher}

 \subsection{Incomplete Verification via Probabilistic Approaches}
 \label{sec:probabilistic-robustness-verification-root}
 
    Besides deterministic verification, one recently emerging branch of studies proposes to add random noise to smooth the models, and thus derive the certified robustness for these \emph{smoothed models}~(See \Cref{def:randomized_classifier}). 
    We call this line of work \emph{probabilistic robustness verification} approaches or \emph{randomized smoothing based} approaches
    since they provide probabilistic robustness guarantees and all existing probabilistic verification approaches are designed for smoothed models.
    Currently, only these verification approaches are scalable enough to certify nontrivial robustness on the large-scale ImageNet dataset.
    
    \begin{definition}[Smoothed Classifier]
        \label{def:randomized_classifier}
        \label{eq:h-smooth}
        Given a smoothing distribution $\mu$ whose support is $\supp(\mu)$ and density at point $\delta$ denoted by $\mu(\delta)$. 
        For a given classifier $F$,  the \emph{smoothed classifier} $F_{\smooth}$ is defined as: 
        \vspace{-0.5em}
        \begin{equation*}
            \small
            F_{\smooth}(x)
            = \argmax_{i \in [C]} \int_{\delta\in \supp(\mu)} \1\left[F(x + \delta)=i\right]\mu(\delta) \dif \delta.
        \end{equation*}
        \vspace{-1.5em}
    \end{definition}
    
    The integral in \Cref{eq:h-smooth} cannot be exactly solved. Thus, instead, Monte-Carlo estimation and hypothesis testing
    \cite{hung2019rank} are used to approximate the exact solution. As a result, the certification is probabilistic rather than deterministic~(\Cref{def:verification-approach}). 
    
    \subsubsection{\textbf{Approaches with Zeroth-Order Information}\texorpdfstring{~\cite{cohen2019certified,Dvijotham2020A,lecuyer2019certified,lee2019tight,li2019certified,salman2019provably,teng2020ell,yang2020randomized,Zhang2020BlackBoxCW}}{}}
    \label{subsec:zeroth-order-approach}
    A majority of verification approaches only use zeroth-order information of the smoothed classifier, i.e., the probabilities $\Pr_{\delta\sim\mu} [F(x_0+\delta)=y]$ for $y\in [C]$ where the clean input is $x_0$, to compute the robustness certification.
    Among these approaches, Neyman-Pearson based approaches are proved to be the tightest~\cite{yang2020randomized}.

    Given clean input $x_0$, when adding noise $\delta$, we suppose the model $F$ predicts true class $y_0$ with probability $P_A := \Pr_{\delta\sim\mu} [F(x_0+\delta) = y_0]$ and runner-up class with 
    $P_B := \max_{y'\in [C]: y'\neq y_0} \Pr_{\delta\sim\mu} [F(x_0+\delta)=y']$.
    High-confidence intervals for $P_A$ and $P_B$ can be obtained with Monte-Carlo sampling. 
    %
    The high-level \textbf{intuition} for Neyman-Pearson based approaches is: if the attacker's perturbed input $x$ is close to $x_0$, the distribution of $x+\delta$ would highly overlap the distribution of $x_0+\delta$ where $\delta\sim\mu$ is the added smoothing noise.
    Therefore, the corresponding $P_A'$ and $P_B'$ for perturbed input $x$ will not change too much from $P_A$ and $P_B$ for clean input $x_0$.
    That means, if there is a sufficient margin between $P_A$ and $P_B$, then $P_A'$ will still be larger than $P_B'$.
    Thus, the smoothed classifier will still predict $y_0$ for perturbed input $x$ according to \Cref{def:randomized_classifier}.
    
    Formally, based on Neyman-Pearson lemma~\cite{neyman1933ix}, one can derive a tight lower bound for $P_A'$ and upper bound for $P_B'$ given $P_A$, $P_B$, and input shift (i.e., $x-x_0$).
    Then, we solve the distance lower bound that guarantees $P_A' > P_B'$ to get robustness certification.

    \textbf{Against $\cL_2$ adversary}, Cohen~et~al~\cite{cohen2019certified} consider Gaussian smoothing and derive a tight $\ell_2$ robustness radius based on Neyman-Pearson lemma.
    \textbf{Against $\cL_1$ adversary},
    Lecuyer~et~al~\cite{lecuyer2019certified} and Teng~et~al~\cite{teng2020ell} consider Laplacian smoothing and derive $\ell_1$ robustness radius.
    \textbf{Against $\cL_\infty$ adversary},
    Yang~et~al~\cite{yang2020randomized} empirically show and theoretically justify that it yields the highest $\cL_\infty$ certified radius by using Gaussian smoothing and transforming Neyman-Pearson-based $\ell_2$ robustness radius to $\ell_\infty$ radius: $r\mapsto \frac{r}{\sqrt d}$ where $d$ is the input dimension.
    However, for dataset where $d$ is large, the certified $\ell_\infty$ radius is small.
    Indeed, certifying robustness against $\ell_\infty$ for high-dimensional input is proven to be intrinsically challenging for zeroth-order information approaches~\cite{blum2020random,kumar2020curse,mohapatra2020rethinking,yang2020randomized}.

    Using zeroth-order information, the robustness verification can also be derived from:
    (1)~differential privacy~(DP) where Gaussian and Laplace mechanisms in DP can induce certified robustness for models smoothed with Gaussian and Laplace distributions~\cite{lecuyer2019certified};
    (2)~Lipschitz bound~\cite{salman2019provably};
    (3)~statistics view~\cite{Dvijotham2020A,li2019certified};
    and (4)~level-set method~\cite{yang2020randomized}.
    These approaches derive looser~\cite{lecuyer2019certified,li2019certified} or equivalently tight~\cite{Dvijotham2020A,lee2019tight,salman2019provably,teng2020ell,Zhang2020BlackBoxCW} robustness certification as Neyman-Pearson based approaches.
    
    \subsubsection{\textbf{Approaches with First-Order Information}\texorpdfstring{~\cite{levine2020tight,mohapatra2020higher}}{}}
    Since zeroth-order information approaches have tightness barriers as discussed before, attempts have been made on querying more information from the smoothed model beyond only $P_A$ and $P_B$.
    One example could be the gradient magnitude information $\|\frac{\partial\Pr_{\delta\sim\mu}[F(x_0+\delta)=y_0]}{\partial x_0}\|_p$ which can be estimated via sampling with high-confidence error interval~\cite{levine2020tight,mohapatra2020higher}.
    When using this first-order information together with $P_A$ and $P_B$, we can derive a tighter robustness verification for smoothed models.
    Currently, the tightness improvements are pronounced against $\ell_1$ adversary but not significant against $\ell_2$ or $\ell_\infty$ adversaries.
    
    \subsubsection{\textbf{Choice of Smoothing Distributions}}
    To achieve satisfactory certified robustness, besides verification approaches, the choice of smoothing distribution is also important for these randomized smoothing based approaches.
    \label{subsec:choice-of-smooth-dist}
    
    In general, for $\ell_2$ adversary, Gaussian smoothing distribution is most commonly used~\cite{cohen2019certified,yang2020randomized}.
    Some work argues that Gaussian may not be the optimal smoothing distribution~\cite{Zhang2020BlackBoxCW} but significantly better alternatives have not been found yet.
    %
    For $\ell_1$ adversary, 
    Yang~et~al~\cite{yang2020randomized} show that uniform distribution is significantly better than others~\cite{lecuyer2019certified,li2019certified,teng2020ell}.
    Recently, a specific non-additive discrete smoothing distribution is proposed~\cite{levine2021improved}, which enables Lipschitz-bound-based \emph{deterministic} certification for smoothed models against $\ell_1$ adversary.
    Since deterministic verification does not need to consider sampling error, it provides better robustness certification than \cite{yang2020randomized}.
    
    The smoothing distributions can control trade-offs between certified robustness and accuracy, where distribution with larger variance can lead to a larger certified radius under the same $P_A$ and $P_B$, but hurts the clean accuracy since input signal is more severely corrupted by noise~\cite{kumar2020curse,mohapatra2020rethinking}.
    
    \begin{practitioner}
        The confidence level of probabilistic verification is usually set to be high~(e.g., $99.9\%$) to maintain the rigor of robustness certification.
        Probabilistic verification approaches are very strong against $\ell_1$ and $\ell_2$ adversaries, being the only type that can provide certification on large-scale datasets~(e.g., ImageNet) and achieve the state-of-the-art on almost all other datasets.
        Moreover, they support arbitrary model architectures since only black-box access to final predictions is required.
        However, they fail to provide robustness certification that is tight enough against the $\ell_\infty$ adversary on ImageNet.
        Furthermore, the smoothing procedure adds additional overhead for inference and requires specific training to ensure good model performance which usually hurts the clean accuracy~\cite{mohapatra2020rethinking}.
    \end{practitioner}
    
    \begin{researcher}
        For verification approaches for smoothed DNNs, promising research directions include: (1)~Tighter verification: though zeroth-order verification fails to certify a large radius against $\ell_\infty$ adversary, other information can be leveraged to tighten it.
        For example, approaches using first-order information are visibly tighter than zeroth-order verification against $\ell_1$ adversary~\cite{mohapatra2020higher}.
        However, there is still a visible gap between the current certified robustness radius and the empirically attackable radius~\cite{hayes2020extensions}, so improvement rooms exist.
        Thus, it would be a promising direction to derive tighter verification by using effective information in addition to zeroth-order information, especially for the challenging $\ell_2$ and $\ell_\infty$ adversaries.
        A recent work~\cite{li2022double} follows this direction and proposes double sampling randomized smoothing that achieves tighter verification against $\ell_2$ adversary.
        In App.~\ref{adxsec:double-sampling-l-inf}, we extend their method and achieves tighter verification against $\ell_\infty$ adversary.
        Researchers can think about what types of information from the smoothed models are useful and how to efficiently obtain them---novel sampling methods are needed.
        (2)~Better smoothing distributions: suitable smoothing distributions have significantly improved $\ell_1$ certified robustness~\cite{levine2021improved,yang2020randomized}. The same degree of improvements may exist against other $\ell_p$ adversaries.
        (3)~Better training approaches for smoothed classifiers.
        Discussion is in \Cref{sec:training-approach}. 
    \end{researcher}
    
    \noindent
    
\section{Robust Training Approaches}
    \label{sec:training-approach}
    
    Normally-trained DNNs are usually non-robust where effective attacks can find adversarial examples with almost 100\% probability~\cite{madry2017towards,papernot2016distillation,szegedy2013intriguing}.
    To achieve high certified robustness, DNNs need to be trained with robust training approaches which aim to improve robustness guarantees, as illustrated in \Cref{fig:categorization-high-level}.
    
    Current robustness verification approaches usually favor certain properties of DNNs to achieve high certified robustness.
    For instance:
    (1)~Branch-and-bound verification~(\Cref{subsec:branch-and-bound}) uses incomplete verification such as linear relaxation to reduce explored branches and boost the certification efficiency so it favors DNNs whose linear relaxations are tight and for other DNNs the certification process is significantly slower~\cite{palma2021scaling,wang2021beta}.
    (2)~Linear relaxation verification~(\Cref{sec:lp-relaxation-intro}) can certify only models whose specific linear relaxations are tight.
    (3)~Lipschitz or curvature verification~(\Cref{sec:lipschitz-intro}) can certify only models where a small Lipschitz or curvature constant can be computed.
    (4)~Verification for smoothed DNNs~(\Cref{sec:probabilistic-robustness-verification-root}) certifies larger radius for models with higher correct-prediction probability under noise.
    These favored properties are not directly promoted by standard training or empirical defenses.
    Therefore, to improve \emph{certified} robustness, robust training approaches are proposed to promote these properties during training.
    
    We divide existing robust training approaches into three categories: regularization-based, relaxation-based, and augmentation-based. 
    We defer the illustration to App.~\ref{adxsec:robust-training-approaches}.
    
    \para{Discussion.}
    Robust training approaches can improve model robustness and at the same time remarkably enhance desired properties of models for corresponding verification approaches.
    Thus, models trained with a robust training approach usually achieve much better certified robustness based on corresponding verification, as reflected by evaluation in \Cref{subsec:benchmark-evaluation}.
    
    Models trained by one robust training approach are often verified to have poor robustness by a mismatched verification approach~(as shown in \Cref{subsec:benchmark-evaluation}). 
    This is because the models do not inherit the desired property of the verification.
    For example, models trained for randomized smoothing based approach can predict well for noisy input but may have many unstable neurons and loose linear relaxations, making them difficult to be verified by complete or linear relaxation verification.
    Models trained for linear relaxation are not specialized for predicting noisy input and are challenging for randomized smoothing based verification~\cite{cohen2019certified}.
    
    As a result, an important research goal is to develop verification that does not heavily rely on specific model properties so it can verify existing robustly trained or empirically defended models.
    Recently, some complete verification approaches~\cite{palma2021scaling,wang2021beta} are shown tractable for verifying empirically defensed~(e.g., PGD adversarially trained~\cite{madry2017towards}) models though they are still limited to small models.
    Thus, proposing more practically efficient complete verification approaches may be a viable path towards this goal.
    Another research goal is to improve certified robustness for a given task, by improving from both the verification side and the robust training side, such as tighter~\cite{tjandraatmadja2020convex} or more training-friendly relaxation~\cite{jovanovic2021certified}, or more effective training methods~\cite{jeong2021smoothmix,shi2021fast,zhang2021boosting}, which we will discuss further in~\Cref{sec:discussion}.

    \vspace{-0.25em}
\section{Benchmark, Leaderboard, and Implications}
    \vspace{-0.25em}
    \label{sec:evaluation}

\input{tables/sota_table}

    In this section, we introduce an open-source toolbox to systematically benchmark $20+$ certifiably robust approaches.
    Based on benchmark results and the leaderboard on representative datasets, we outline practical implications for deploying certifiably robust approaches for DNNs.

    
    \vspace{-0.5em}
    \subsection{Benchmark Evaluation}
        \label{subsec:benchmark-evaluation}
    \vspace{-0.3em}
    We present the following evaluation:
    (1)~for representative \textbf{deterministic verification} approaches, we compare their certified robustness over a diverse set of trained models of different scales;
    (2)~for representative \textbf{probabilistic verification} approaches and their corresponding robust training approaches, we compare the best certified robustness they jointly achieve.
    We do such separation because deterministic and probabilistic certificates have different semantics and their supported system models are different.
    The evaluation is made possible by our open-source unified toolbox---a first toolkit integrating a wide range of verification approaches.
    
    
    We present the findings here and introduce the experiment protocol and representative results in App.~\ref{adxsec:benchmark-eval-detail}.
    Full results are on our benchmark website: \textit{\url{https://sokcertifiedrobustness.github.io}}.
    In the evaluation, we use \textit{certified accuracy} to measure certified robustness, which is the fraction of test set samples verified to be robust against the corresponding $(\ell_p,\epsilon)$-adversary.

    \subsubsection{Findings from Comparing Deterministic Verification Approaches}
        (1)~On relatively small models, complete verification approaches can effectively verify robustness, thus they are the best choice.
        (2)~On larger models, usually linear relaxation based verification approaches perform the best since the complete verification approaches are too slow and other approaches are too loose, yielding almost $0\%$ certified accuracy.
        However, linear relaxation based verification still cannot handle large DNNs and they are still too loose compared with the upper bound provided by PGD attack.
        (3)~On robustly trained models, if the robust training approach is \t{CROWN-IBP} which is tailored for \t{IBP} and \t{CROWN}~(two linear relaxation verification approaches), \t{IBP} and \t{CROWN} can certify high certified accuracy while others fail to certify.
        Indeed, robust training approaches can usually boost the certified accuracy but the models must be verified with corresponding verification approaches as discussed in \Cref{sec:training-approach}.
        (4)~SDP approaches usually take too long and thus are less practical.

    \subsubsection{Findings from Comparing Probabilistic Verification Approaches}
        (1)~For both $\cL_1$ and $\cL_2$ adversaries, Neyman-Pearson based verification achieves the highest certified robustness.
        (2)~Robust training approaches effectively enhance the models' certified robustness. Among these existing robust training approaches, adversarial training usually achieves the best performance. 
        (3)~The choice of smoothing distribution can greatly affect the certified accuracy. Under $\cL_1$ adversary, the superior result is achieved by uniform smoothing distribution.
        (4)~For probabilistic verification approaches, certifying robustness under $\ell_\infty$ adversary is challenging, and would become more challenging when the data dimension increases, which coincides with theory~\cite{blum2020random,kumar2020curse,yang2020randomized}.
    
    \vspace{-0.5em}
    \subsection{Leaderboard on Certified Robustness}
    \vspace{-0.5em}
        \label{sec:leaderboard}
        
        \emph{What is the state-of-the-art certified accuracy achieved on representative datasets?}
        \Cref{tab:leaderboard} shows a leaderboard of certified accuracy under different settings from peer-reviewed publications till April 1, 2023. 
        The high certified accuracy is jointly achieved by robust training (shown by reference bracket) and verification (shown by name).

        As we can see, much progress has been made in the certified robustness field in recent years.
        On MNIST, the certified accuracy against $\cL_\infty$ adversary with $0.3$ radius has reached over $94\%$.
        This is remarkable since the limit is $0.5$ radius where any input image can be perturbed to indistinguishable half-gray.
        This is achieved by robust training for linear relexation~\cite{lyu2021towards}. 
        On more challenging CIFAR-10 and ImageNet datasets, however, certified accuracy is still low.
        On CIFAR-10, against $\cL_\infty$ adversary with $2/255$ radius, the certified accuracy is only around $68\%$~\cite{salman2019provably}; with radius $8/255$, it is only $40.39\%$~\cite{zhang2022rethinking}.
        These are far from state-of-the-art $90\%+$ clean accuracy or $65\%+$ accuracy under strong attacks~\cite{croce2020robustbench}.
        On ImageNet, only approaches for smoothed DNNs can provide non-zero robust accuracy: $42.2\%$ under $\cL_2$ radius $2.0$~\cite{xiao2023densepure}; and around $38\%$ under $\cL_\infty$ radius $1/255$~\cite{salman2019provably}.

    \begin{figure}[!t]
        \centering
        \vspace{-0.5em}
        \includegraphics[width=\linewidth]{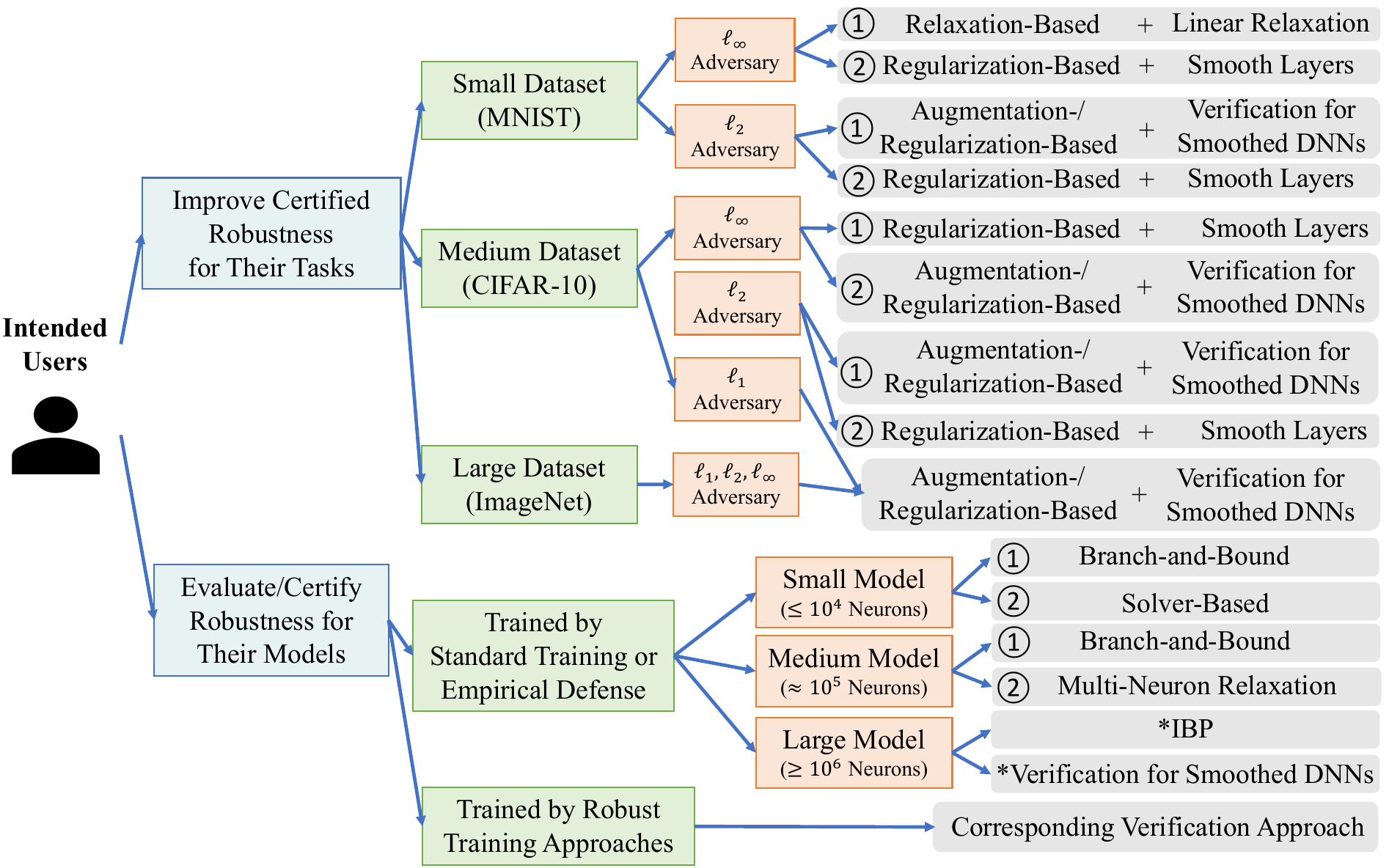}
        \vspace{-2em}
        \caption{Practical implications for users to select certifiably robust approaches.
        Gray boxes depict suitable ``(verification) + (robust training)'' approach combinations or verification approaches for given scenarios. If exist, ``\textcircled{1}'' and ``\textcircled{2}'' label the most and runner-up suitable ones.  Details in \Cref{subsec:guideline}.}
        \label{fig:user-guidance}
        \vspace{-1.5em}
    \end{figure}
    
    \subsection{Practical Implications}
        \label{subsec:guideline}
        \vspace{-0.5em}
        
        \textit{In practice, what are the most suitable certifiably robust approaches for users to deploy?}
        Based on the benchmark results and the leaderboard, we present practical implications in \Cref{fig:user-guidance}, where we envision two scenarios: 1)  users want to improve certified robustness for their tasks at hand; 2) users want to evaluate or certify the robustness of given models.
        
        When users want to improve certified robustness for their tasks, they need to achieve this by choosing a robust training approach and certifying the robustness with the corresponding verification approach as discussed in \Cref{sec:training-approach}.
        The upper part of \Cref{fig:user-guidance} shows the recommended combinations of verification and robust training approaches in light gray boxes.
        Depending on the dataset size and the type of $\ell_p$ adversary to defend against, we recommend the corresponding approach combinations which achieve high certified accuracy in practice based on our leaderboard~(\Cref{tab:leaderboard}).
        When multiple choices are available, we show the top-2 choices and label them with ``\textcircled{1}'' and ``\textcircled{2}'' respectively.
        In summary, linear-relaxation-based verification is suitable only on small datasets against $\ell_\infty$ adversary, Lipschitz-based~(including smooth layers, Lipschitz, and general Lipschitz) verification is suitable on small and medium datasets, and smoothed DNNs~(including Neyman-Pearson-based and differential-privacy-based) are suitable on medium and large datasets.
        In particular, on large datasets like ImageNet, only approaches for smoothed DNNs can provide robustness certification at the current stage.
        
        When users want to evaluate or certify the robustness of certain models, they need to choose a suitable verification approach.
        Inspired by our benchmark, we present the implications in the lower part of \Cref{fig:user-guidance}.
        For small and medium models trained by standard training or empirical defenses, the branch-and-bound based complete verification~\cite{wang2021beta,ferrari2021complete} and multi-neuron relaxation verification~\cite{palma2021scaling} can certify the robustness efficiently.
        Specifically, for small models, the solver-based~(concretely, MILP-based) verification approaches can certify good robustness.
        But for large models, none of these methods can finish in a feasible time~(one day per input).
        Therefore, we must use more efficient but loose verification such as IBP~(\Cref{sec:linear-inequality-propagation-intro}) and verification approaches for smoothed DNNs~(\Cref{sec:probabilistic-robustness-verification-root}), which usually yield trivial certified robustness radius and it is an active research area to make tighter verification approaches scalable for these large models. 
        In addition, we find that the ranking of empirical defenses for small/medium models based on certified robustness is consistent with that evaluated by strong empirical attacks such as  PGD~\cite{dathathri2020enabling,wang2021beta,wu2021crop}.
        If models are trained by robust training approaches, as discussed in \Cref{sec:training-approach}, using the corresponding verification approaches targeted by the training approach would be the best choice.

\section{Extensions and Applications}
    \label{sec:extensions}


    The methodologies derived from certifiably robust DNNs have recently been applied to much broader areas.
    
    \para{Extensions to other threat models.}
    Though certifiably robust approaches mainly focus on the $(\ell_p,\epsilon)$-adversary, extending the related techniques to other threat models has drawn much attention.
    (1)~\textbf{Local evasion attacks}:
    in local evasion attacks, the adversary slightly perturbs the in-distribution data to mislead the model.
    Our $(\ell_p,\epsilon)$-adversary adds pixelwise perturbations bounded by $\ell_p$ norm within $\epsilon$.
    Now we elaborate on some other effective local evasion attacks.
    (a)~\emph{Semantic adversary} picks an arbitrary but bounded transformation parameter, such as rotation angle, and applies the transformation to perturb the input~\cite{engstrom2017rotation,pei2017deepxplore,xiao2018spatially}.
    Neyman-Pearson approaches and linear relaxation approaches can be extended to provide certification~\cite{fischer2020certification,li2020provable,pautov2021cc,hao2022gsmooth}.
    The core methodology is to split the low-dimensional parameter space into tiny intervals and then bound the input changes in each interval.
    (b)~\emph{Generative model based adversary} uses generative models such as GAN~\cite{goodfellow2014generative} to generate input perturbation.
    Similar to certification against the semantic adversary, Neyman-Pearson approaches and linear relaxation approaches can be extended to provide certification against this adversary~\cite{mirman2021robustness,wong2020learning}.
    (c)~\emph{$\ell_0$ adversary} picks a bounded number of pixels to arbitrarily change and \emph{patch adversary} picks a region of pixels with a bounded area to arbitrarily change.
    To defend against $\ell_0$ adversary, Neyman-Pearson approaches can be deployed~\cite{lee2019tight,levine2020robustness,jia2022almost}.
    To defend against patch adversary, the core idea of Neyman-Pearson approaches, prediction aggregation on several noisy inputs which are patched inputs here, is leveraged to develop customized certification and corresponding training approaches. 
    Starting from direct prediction aggregation~\cite{alex2020derandomized}, some recent certified defenses exploit or design model architectures and inference procedures with self-aggregation property, such as DNNs with small localized receptive fields~\cite{xiang2021patchguard}, importance-score-based pruning~\cite{han2021scalecert},  vision transformers~\cite{salman2021certified}, and two-round patch-masking~\cite{xiang2022patchcleanser}, to improve the efficiency and tightness of robustness certification.
    These approaches~\cite{alex2020derandomized,xiang2021patchguard,han2021scalecert,salman2021certified,xiang2022patchcleanser} can provide robustness guarantees on the large-scale ImageNet dataset.
    (2)~\textbf{Distributional evasion attacks}:
    in distributional evasion attacks, the attacker shifts the whole test data distribution within some bounded distance to maximize the expected loss.
    This threat model can be used to characterize the out-of-distribution generalization ability of ML models~\cite{shen2021towards}.
    The certification under this threat model is an upper bound of the expected loss, which can be derived from duality under Lipschitz and curvature assumptions~\cite{sinha2018certifying}
    or from extensions of
    Neyman-Pearson approaches~\cite{kumar2022certifying,weber2022certifying}.
    (3)~\textbf{Global evasion attacks}:
    global evasion attacks can perturb any valid input example to mislead the model, whereas local evasion attacks can only perturb in-distribution data.
    Thus, the robustness against global evasion attacks means that the robustness property holds for the whole input domain.
    An example of a robustness property is that for any high-confident prediction, small perturbations cannot change the predicted label~\cite{globally2021leino}.
    In the security domain, Chen~et~al~\cite{chen2021learning} recently proposed several domain-specific robustness properties such as requiring all low-cost features to be robust.
    To verify these properties, they propose a specific solver-based verification~(\Cref{subsec:solver-based}) to verify \textit{logic ensemble models}, and then use the found adversarial example as an augmentation for robust training.
    The verification and robust training \textit{for DNNs} against global evasion attacks can be a promising direction.
    (4)~\textbf{Training-time attacks}:
    training-time attacks can manipulate some training data to reduce the trained model's performance or inject some backdoors.
    Against this threat model, verification approaches extended from Neyman-Pearson can provide robustness certification~\cite{levine2020deep,rosenfeld2020certified,weber2020rab,wu2022copa}.

    \para{Extensions to diverse types of system models.}
        There are efforts on generalizing existing DNN verification approaches to deal with more types of system models.
        For example: 
        (1)~Some approaches that are designed for feed-forward ReLU networks, such as linear relaxation based approaches, have been extended to support general DNNs~\cite{zhang2018efficient,wong2018scaling,singh2019abstract}, recurrent networks~\cite{du2021cert,ko2019popqorn,ryou2021scalable}, transformers~\cite{bonaert2021fast,Shi2020Robustness}, generative models~\cite{mirman2020robustness}, and model ensembles~\cite{zhang2019enhancing}.
        The main methodology is to derive the corresponding linear bounds for activation functions or attention mechanisms in these system models.
        Some complete verification approaches, e.g., branch-and-bound based ones~\cite{wang2021beta}, also support general DNNs.
        However, these complete verification approaches become incomplete when applied on general DNNs. 
        (2)~Verification approaches for Lipschitz-bounded networks and non-ReLU networks have not been generalized to other system models yet.
        (3)~Verification approaches for smoothed DNNs typically need access to only the final prediction label, so they are applicable to any classification models. 
        However, the model must follow the corresponding smoothing-based inference protocol.
        (4)~There are also verification approaches for decision trees~\cite{andriushchenko2019provably,chen2019robustness,wang2020lp}, decision stumps~\cite{wang2020lp}, nearest prototype classifiers~\cite{voravcek2022provably}, and logic ensembles~\cite{chen2021learning}.
        However, there is no verification and robust training approach that supports all these system models yet.
        This is because verification and robust training approaches need to exploit properties~(piecewise linearity, Lipschitz bound, smoothness, etc) of specific system models to achieve certified robustness.

    \para{Certified robustness for concrete applications.}
    %
        Beyond the classification task, the discussed methodologies, such as linear relaxation and Neyman-Pearson approaches, have been extended to certify DNNs in many concrete applications.
        In natural language processing, extensions include certification for recurrent neural networks against embedding perturbations~\cite{du2021cert,ko2019popqorn,ryou2021scalable}, word substitutions~\cite{Jia2019CertifiedRT}, and word transformations~\cite{ye2020safer,zhang2020robustness,zhang2021lstm}.
        Extensions have also been studied for object detection~\cite{chiang2020detection}, segmentation~\cite{fischer2021scalable}, and point cloud models~\cite{fischer2021scalable,chu2022tpc,liu2021pointguard} in computer vision, and speech recognition~\cite{fischer2020certification,olivier2021sequential}.
        Verification and robust training approaches have also been proposed for reinforcement learning~\cite{kumar2021policy,lutjens2019certified,wang2019verification,wu2021crop,wu2022copa}.
\section{Insights, Challenges, and Future Directions}
    \label{sec:discussion}
    
    In this section, we summarize characteristics, strengths, limitations, and fundamental connections among certifiably robust approaches, then discuss barriers, main challenges, and future directions for DNN certification.
    
    \para{A unified view: characteristics, strengths, limitations, and connections of certifiably robust approaches.}
    To reveal the fundamental connections, we adopt a unified view of robustness verification: all existing verification approaches provide an abstraction of given DNN models to verify the robustness.
    For example, the branch-and-bound verification views the model as the union of several sub-domains where the model output in each domain can be bounded, e.g., by linear inequalities.
    The branching process is essentially refining the abstraction by splitting sub-domains whose current abstractions are not precise enough.
    The linear relaxation based verification uses some linear constraints to abstract the possible behavior of the model in the whole perturbation region.
    The probabilistic verification uses the queried information, such as zeroth-order information, to abstract the model behavior.
    This view is closely related to the concept of abstract interpretation in traditional program analysis~\cite{cousot1977abstract}.
    Therefore, the scalability and tightness trade-off of verification mentioned in \Cref{sec:definition-robust-verification-provable-defense} is essentially the inherent trade-off between preciseness and efficiency of abstraction:
    more precise abstraction enables tighter robustness certification, whereas has higher time and space complexity.
    Thus, for a model that is not specifically trained, the most suitable verification approach is the most precise one that can be computed for this model size.
    Concrete approach selection guidelines are in \Cref{subsec:guideline}.
    We note that, under this unified view,
    the favored properties of each verification~(listed in \Cref{sec:training-approach}) are tight conditions of the corresponding abstraction domain.
    Thus, robust training approaches that promote these properties can boost verification tightness for the model to improve certified robustness.
    More concrete strengths and limitations of each verification are discussed in ``practical implications'' and ``research implications'' boxes in \Cref{sec:verification-approaches}.
    
    \para{Challenges and barriers.}
    Although there has been remarkable progress towards certifiably robust DNNs, scalability and tightness challenges persist.
    For example: (1)~Complete verification is $\NP$-complete~\cite{katz2017reluplex,weng2018towards}.
    (2)~Multi-neuron based linear relaxation needs exponential number of constraints~\cite{tjandraatmadja2020convex}.
    (3)~Probabilistic certification based on zeroth-order information cannot certify high robustness against $\ell_\infty$ adversary for real-world high-dimensional inputs~\cite{blum2020random,kumar2020curse,mohapatra2020rethinking,yang2020randomized}.
    These theoretical barriers are intrinsic challenges for further improvements in these verification approaches.
    There are also practical issues to solve, such as guaranteeing verification soundness under floating-point arithmetic~\cite{jia2021exploiting,zombori2020fooling,voravcek2022sound} and safeguarding robust training against training-time attacks~\cite{mehra2021robust}.
    
    \para{Future directions.}
    Despite the challenges and barriers, there are also several potential future directions:
    (1)~\textbf{Scalable and tight verification:}
    There are still hopes for more scalable and tighter verification for DNNs \emph{in practice} despite theoretical barriers.
    For example, good heuristics have boosted complete verification to handle DNNs with over $10^5$ neurons~\cite{wang2021beta}.
    It is promising to explore other better heuristics.
    For instance, a recent work~\cite{ferrari2021complete} improves the complete verification by proposing better bounding heuristics based on multi-neuron relaxation.
    For SDP verification, better formulation and solvers can lead to better verification~\cite{batten2021efficient,dathathri2020enabling}.
    For smoothed DNNs, although only using zeroth-order information cannot certify high robustness against $\ell_\infty$ adversary, this barrier may be circumvented by leveraging more information as in \cite{levine2020tight,mohapatra2020higher} which improve $\ell_1$ and $\ell_\infty$ certification tightness; or leveraging non-additive smoothing distribution as in \cite{levine2021improved} which improves $\ell_1$ certification tightness.
    More details are discussed in \Cref{sec:probabilistic-robustness-verification-root}.
    (2)~\textbf{Effective robust training with theoretical understanding:}
    Unlike verification where theoretical barriers exist, robust training can empirically boost the certified robustness without known theoretical limitations.
    Indeed, even the empirically loose interval relaxations~(see \Cref{sec:ibp-intro}) are universal approximators~\cite{Baader2020Universal,wang2020interval} and achieve training convergence~(under some assumptions)~\cite{wang2022on}, which implies that with effective and generalizable robust training the certified accuracy could be on par with benign accuracy.
    However, theoretical understanding of robust training, such as why robust training generalizes, is still lacking~\cite{jovanovic2021certified}.
    Recent work shows that when an efficient complete verification approach exists, generalizable robust training is achievable~\cite{ashtiani2020black}.
    Extending this result to broader scenarios, e.g., the generalization of robust training with \emph{incomplete} verification, would significantly advance our understanding of ML robustness.
     (3)~\textbf{Design certifiably robust DNN architectures}: Based on the model properties required for different verification approaches, it is promising to design novel DNN architectures to further improve the certified robustness. In addition, it is also possible to design sparse DNNs following the model compression literature~\cite{sehwag2020hydra} to achieve efficient and certifiably robust models.
    (4)~\textbf{Certification for other ML utilities:}
    Techniques of certified robustness can be extended to certify other ML utilities such as fairness~\cite{ruoss2020learning,urban2020perfectly,kang2022certifying} and generalization~\cite{kumar2022certifying,weber2022certifying}.
    It is an emerging trend to provide certification for generic ML utilities, such as model bias, toxicity, and model unlearning, or train models to achieve such certifications~\cite{bourtoule2021machine}.
     (5)~\textbf{Certification for different ML models:} Current robustness certification mainly focuses on classification models, and it would be critical to extend such certification to other ML models, such as reinforcement learning, federated learning, and large language models, which have demonstrated their real-world usage in safety-critical domains.
     (6)~\textbf{Integrate domain knowledge and logic reasoning ability into ML to improve certified robustness:} It has been shown that joint inference with knowledge rules can improve model benign accuracy~\cite{qu2019probabilistic,xie2019embedding,zhang2019efficient}, and therefore it would be promising to integrate domain knowledge, causal analysis, and security rules into ML pipeline to further improve and tighten its end-to-end certified robustness.
    (7)~\textbf{Bring certified robustness to real-world applications:}
    Besides achieving higher certified robustness on standard benchmarks, we believe that adaptation of verification and robust training approaches for real-world applications is also critical.
    For example, security threats are found on DNNs in autonomous vehicles~\cite{hallyburton2021security,cao2021invisible} which may lead to severe consequences~\cite{selfdrivingkills,pei2017deepxplore}.
    Designing a certifiably robust autonomous driving system would be an important, timely, and promising direction.

\section{Conclusions}
    We presented an SoK for certifiably robust approaches for DNNs, including both robustness verification approaches and robust training approaches.
    We show characteristics, strengths, limitations, and fundamental connections among these approaches.
    Our discussion summarizes the current research status both theoretically and empirically, reveals limitations, and highlights future directions.

\ifnum\arxiv=1
\section*{Acknowledgment}
    We would like to thank Xiangyu Qi for conducting the benchmark evaluation on some probabilistic verification approaches for smoothed DNNs.
    We thank Dr. Ce Zhang, Dr. Sasa Misailovic, and Dr. Gagandeep Singh for their thoughtful feedback.
    We also thank the support of NSF grant No.1910100, NSF CNS 2046726, C3 AI,  the Alfred P. Sloan Foundation, and the AWS Research Awards.
\fi


%% file: tables/property-table.tex
\begin{table*}[!t]
    \centering
    \vspace{-0.8em}
    \caption{Properties and references of robustness verification approaches. 
    Notations are explained in \Cref{sec:main-taxonomy}.
    }
    
    \vspace{-1em}
    \resizebox{0.95\textwidth}{!}{
    \begin{tabular}{cccccccccccccc}
        \toprule
        Complete/ & Deterministic/ & \multicolumn{2}{c}{\multirow{2}{*}{System Model}} & \multicolumn{3}{c}{\multirow{2}{*}{Robustness Verification Approaches}} & \multicolumn{3}{c}{Supported $\ell_p$} &
        \multicolumn{2}{c}{Scalability} & \multirow{2}{*}{Tightness} & \multirow{2}{*}{References} \\
        \cline{8-10}
        Incomplete & Probabilistic & & & & & & $\ell_\infty$ & $\ell_2$ & $\ell_1$ & (Scale up to) & (Complexity) \\
        \midrule
        \arrayrulecolor{tlegray}
        
        \multirow{5}{*}{Complete} & \multirow{5}{*}{Deterministic} & \multicolumn{2}{c}{\multirow{5}{*}{\shortstack{for Feed-\\Forward\\ ReLU Nets}}} & \multirow{2}{*}{Solver-Based} & \multicolumn{2}{c}{SMT-Based} & \checkmark & \checkmark & \checkmark & MNIST & $O(2^{lw})$ & Complete & \cite{pulina2010abstraction,pulina2012challenging} \\ \cline{6-14}
        & & & & & \multicolumn{2}{c}{MILP-Based} & \checkmark & & & CIFAR-10 & $O(2^{lw})$ & Complete & \cite{cheng2017maximum,dutta2017output,lomuscio2017approach,tjeng2018evaluating} \\
        \cline{5-14}
        & & & & \multicolumn{3}{c}{Extended Simplex Method} & \checkmark & & & MNIST & $O(2^{lw})$ & Complete & \cite{katz2017reluplex,katz2019marabou} \\
        \cline{5-14}
        & & & & \multicolumn{3}{c}{\multirow{2}{*}{Branch-and-Bound}} & \multirow{2}{*}{\checkmark} & \multirow{2}{*}{\checkmark} & & \multirow{2}{*}{CIFAR-10} & \multirow{2}{*}{$O(2^{lw})$} & \multirow{2}{*}{Complete}  & \cite{wang2018efficient,bak2020improved,bunel2019branch,bunel2018unified,de2021improved,ehlers2017formal,ferrari2021complete,fromherz2021fast,gehr2018ai2} \\
        & & & & & & & & & & & & & \cite{jordan2019provable,Lu2020Neural,wang2018formal,wang2021beta,xu2021fast,zhang2022general} \\
        \hline
        
        \multirow{17}{*}{Incomplete} & \multirow{11}{*}{Deterministic} & \multicolumn{2}{c}{\multirow{8}{*}{\shortstack{for General\\ DNNs$^1$}}} & \multirow{6}{*}{\shortstack{Linear\\ Relaxation}} & \multicolumn{2}{c}{Linear Programming (LP)} & \checkmark & (\checkmark) & (\checkmark) & CIFAR-10 & $O(\poly(l, w))$ & ${T_5}^2$ & \cite{salman2019convex,weng2018towards} \\
        \cline{6-14}
        & & & & & \multirow{4}{*}{\shortstack{Linear\\ Inequality}} & Interval & \checkmark & (\checkmark) & (\checkmark) & Tiny ImageNet & $O(lw^2)$ & $T_2$ & \cite{gowal2019scalable} \\
        \cline{7-14}
        & & & & & & Polyhedra & \checkmark & (\checkmark) & (\checkmark) & Tiny ImageNet & $O(lw^3)$ & ${T_4}^2$ & \cite{lyu2020fastened,lyu2021towards,singh2019abstract,weng2018towards,xu2020automatic,zhang2018efficient} \\
        \cline{7-14}
        & & & & & & Zonotope & \checkmark & (\checkmark) & (\checkmark) & Tiny ImageNet & $O(lw^3)$ & ${T_3}^2$ & \cite{anderson2019optimization, mirman2018differentiable,singh2018fast,singh2018robustness} \\
        \cline{7-14}
        & & & & & & Duality & \checkmark & (\checkmark) & (\checkmark) & Tiny ImageNet & $O(lw^3)$ & ${T_4}^2$ & \cite{dvijotham2018training,dvijotham2018dual,wong2018provable,wong2018scaling} \\
        \cline{6-14}
        & & & & & \multicolumn{2}{c}{Multi-Neuron Relaxation} & \checkmark & (\checkmark) & (\checkmark) & CIFAR-10  & $O(lw^3)$ - $O(2^{lw})^6$ & $T_7$ & \cite{muller2022prima,palma2021scaling,singh2019beyond,tjandraatmadja2020convex} \\
        \cline{5-14}
        
        & & & & \multicolumn{3}{c}{Semidefinite Programming (SDP)} & $\checkmark$ & & & CIFAR-10 & $O(\poly(l, w))$ & $T_6$ & \cite{dathathri2020enabling,dvijotham2019efficient,fazlyab2019safety,raghunathan2018certified,raghunathan2018semidefinite} \\
        \cline{5-14}
        
        & & & & \multirow{2}{*}{\shortstack{Lipschitz}} & \multicolumn{2}{c}{General Lipschitz} & & \checkmark & & Tiny ImageNet & $O(lw^2)$ & ${T_1}^3$ & \cite{lee2020lipschitz,globally2021leino,singla2021fantastic,szegedy2013intriguing,tsuzuku2018lipschitz,weng2018towards,zhang2019recurjac} \\
        \cline{3-4} \cline{6-14}
        & & \multicolumn{2}{c}{for Lip-Bounded Nets} & & \multicolumn{2}{c}{Smooth Layers} & $\checkmark$ & $\checkmark$ & & Tiny ImageNet & $O(lw^2)$ & $^3$ & \cite{hein2017formal,li2019preventing,singla2021householder,trockman2021orthogonalizing,zhang2021towards} \\
        \cline{3-14}
        & & \multicolumn{2}{c}{for Non-ReLU Nets$\,^4$} & \multicolumn{3}{c}{Curvature} & & $\checkmark$ &  & CIFAR-10 & $O(lw^3)$ & $\,^4$ & \cite{singla2020second} \\
        \cline{3-14}
        
        & & \multicolumn{2}{c}{for Smoothed DNNs} & Zeroth Order & \multicolumn{2}{c}{Lipschitz} & & & $\checkmark$ & ImageNet & $O(Slw^2)$ & $\,^5$ & \cite{levine2021improved} \\
        \cline{2-14}
        
         & \multirow{6}{*}{Probabilistic} & \multicolumn{2}{c}{\multirow{6}{*}{for Smoothed DNNs}} & \multirow{5}{*}{Zeroth Order} & \multicolumn{2}{c}{Differential Privacy Inspired} & & \checkmark & \checkmark & ImageNet & $O(Slw^2)$ & $ST_1$ & \cite{lecuyer2019certified} \\
        \cline{6-14}
        & & & & & \multicolumn{2}{c}{Divergence Based} & & \checkmark & \checkmark & ImageNet & $O(Slw^2)$ & $ST_2$ & \cite{li2019certified,Dvijotham2020A} \\
        \cline{6-14}
        & & & & & \multicolumn{2}{c}{Neyman Pearson} & & \checkmark & & ImageNet & $O(Slw^2)$ & $ST_3$ & \cite{cohen2019certified} \\
        \cline{6-14}
        & & & & & \multicolumn{2}{c}{Level-Set Analysis} & (\checkmark) & \checkmark & \checkmark & ImageNet & $O(Slw^2)$ & $ST_3$ & \cite{teng2020ell,yang2020randomized,Zhang2020BlackBoxCW} \\
        \cline{6-14}
        & & & & & \multicolumn{2}{c}{Lipschitz} & (\checkmark) & \checkmark & & ImageNet & $O(Slw^2)$ & $ST_3$ & \cite{awasthi2020adversarial,salman2019provably} \\
        \cline{5-14}
        & & & & \multicolumn{3}{c}{First Order} & (\checkmark) & \checkmark & \checkmark & ImageNet & $O(Slw^2)$ & $ST_4$ & \cite{levine2020tight,mohapatra2020higher} \\
        
        
        
        
        
        \arrayrulecolor{black}
        \bottomrule
    \end{tabular}
    }
        
        {
            \flushleft \small
            \begin{minipage}{\textwidth}
                \vspace{-0.6em}
                \scriptsize
                
                1. Typical approaches mainly support feed-forward ReLU networks, but extensions to general DNNs are available~\cite{boopathy2019cnn,Shi2020Robustness,zhang2018efficient}, which are discussed in \Cref{sec:extensions}.\\ 
                2. Tightness depends on intermediate layer bounds. If they share the same intermediate layer bounds, the tightness order is Zonotope $<$ Polyhedra = Duality $<$ LP~\cite{salman2019convex}.\\
                3. Lipschitz bound is loose for typical DNNs, but can be tight for specially regularized DNNs which have small Lipschitz bounds.\\
                4. Only available for networks whose activation functions have nonzero second-order derivatives, which exclude ReLU networks. Thus, tightness is incomparable with others.\\
                5. The approach is designed for some specific smoothing distributions that are not supported by other smoothed DNN oriented approaches. \\
                6. Tunable time complexity dependent on the upper limit of number of linear constraints.
            \end{minipage}
        }
        \vspace{-2.5em}
        
    \label{tab:property-table}
\end{table*}

%% file: tables/sota_table.tex
\begin{table*}[!ht]
    \centering
    \caption{Leaderboard: Top-$5$ certified accuracy under each setting achieved by corresponding robust training (shown by reference bracket) and verification (shown by name). $\epsilon$ denotes the attack radius. 
    $^*$ indicates certified accuracy by probabilistic verification, otherwise by deterministic verification.
     ``Nym.-Prsn.'' means Neyman-Pearson-based or equivalently tight verification. 
    }
    \vspace{-1.3em}
    \resizebox{\linewidth}{!}{

    \begin{tabular}{c|cc|cc|cc|cc|cc|cc}
        \toprule
        & \multicolumn{4}{c|}{$\ell_\infty$ Adversary} & \multicolumn{4}{c|}{$\ell_2$ Adversary} & \multicolumn{4}{c}{$\ell_1$ Adversary} \\
        \hline\hline
        \multirow{6}{*}{MNIST} & \multicolumn{2}{c|}{$\epsilon=0.1$} & \multicolumn{2}{c|}{$\epsilon=0.3$} & \multicolumn{2}{c|}{$\epsilon=0.5$} & \multicolumn{2}{c|}{$\epsilon=1.58$} & \multicolumn{4}{c}{\multirow{6}{*}{\shortstack{Existing approaches for $\ell_1$ are all randomized-smoothing-based,\\ which are generally evaluated on CIFAR-10 and ImageNet datasets.}}} \\
        \cline{2-9}
        & $\mathbf{98.22\%}$ & \cite{mueller2023certified}, Interval 
        & $\mathbf{94.02\%}$ & \cite{lyu2021towards}, Polyhedra
        & $\mathbf{98.2\%}^*$ & \cite{jeong2021smoothmix}, Nym.-Prsn.
        & $\mathbf{70.7\%}^*$ & \cite{jeong2021smoothmix}, Nym.-Prsn.($\epsilon=1.75$) 
        & \\

        & $98.14\%$ & \cite{zhang2022rethinking}, Smooth Layers
        & $93.40\%$ & \cite{zhang2022rethinking}, Smooth Layers
        & $98.0\%^*$ & \cite{jeong2020consistency}, Nym.-Prsn.
        & $70.5\%^*$ & \cite{jeong2020consistency}, Nym.-Prsn.($\epsilon=1.75$) 
        &  \\

        & $97.95\%$ & \cite{shi2021fast}, Interval
        & $93.40\%$ & \cite{mueller2023certified}, Interval
        & $78.45\%$ & \cite{singla2020second}, Curvature
        & $69.79\%$ & \cite{singla2020second}, Curvature
        &  \\
        
        & $97.95\%$ & \cite{zhang2021boosting}, Smooth Layers
        & $93.20\%$ & \cite{zhang2021boosting}, Smooth Layers
        & &
        & $69.0\%^*$ & \cite{li2019certified}, Divergence Based 
        &  \\
        
        & $97.91\%$ & \cite{li2019robustra}, Duality
        & $93.10\%$ & \cite{shi2021fast}, Interval 
        & &
        & $62.8\%$ & \cite{globally2021leino}, General Lipschitz 
        &  \\
        
        \hline
        
        \multirow{6}{*}{CIFAR-10} & \multicolumn{2}{c|}{$\epsilon=2/255$} & \multicolumn{2}{c|}{$\epsilon=8/255$} & \multicolumn{2}{c|}{$\epsilon=36/255$} & \multicolumn{2}{c|}{$\epsilon=0.25$} & \multicolumn{2}{c|}{$\epsilon=1.0$} & \multicolumn{2}{c}{$\epsilon=2.0$} \\
        \cline{2-13}
        & $\mathbf{68.2\%}^*$ & \cite{salman2019provably}, Nym.-Prsn.
        & $\mathbf{40.39\%}$ & \cite{zhang2022rethinking}, Smooth Layers
        & $\mathbf{65.6\%}^*$ & \cite{li2019certified}, Divergence Based
        & $\mathbf{81\%}^*$ & \cite{salman2019provably}, Nym.-Prsn. 
        & $\mathbf{63.07\%}$ & \cite{levine2021improved}, Lipschitz 
        & $\mathbf{51.33\%}$ & \cite{levine2021improved}, Lipschitz \\
        
        & $63.8\%^*$ & \cite{carmon2019unlabeled}, Nym.-Prsn.
        & $40.06\%$ & \cite{zhang2021boosting}, Smooth Layers
        & $64.49\%$ & \cite{xu2022lot}, Smooth Layers
        & $79.3\%^*$ & \cite{carlini2023certified}, Nym.-Prsn.
        & $63\%^*$ & \cite{yang2020randomized}, Nym.-Prsn. 
        & $48\%^*$ & \cite{yang2020randomized}, Nym.-Prsn. \\

        & $62.84\%$ & \cite{mueller2023certified}, Interval
        & $35.42\%$ & \cite{zhang2021towards}, Smooth Layers 
        & $62.96\%$ & \cite{singla2021householder}, Smooth Layers
        & $76.6\%^*$ & \cite{xiao2023densepure}, Nym.-Prsn. 
        & $39\%^*$ & \cite{teng2020ell}, Nym.-Prsn.
        & $17\%^*$ & \cite{Zhang2020BlackBoxCW}, Nym.-Prsn.\\
        
        & $60.5\%$ & \cite{Balunovic2020Adversarial}, Polyhedra
        & $35.13\%$ & \cite{mueller2023certified}, Interval
        & $59.16\%$ & \cite{trockman2021orthogonalizing}, Smooth Layers
        & $72\%^*$ & \cite{carmon2019unlabeled}, Nym.-Prsn. 
        & $34\%^*$ & \cite{Zhang2020BlackBoxCW}, Nym.-Prsn.
        & $16\%^*$ & \cite{teng2020ell}, Nym.-Prsn. \\

        & $56.94\%$ & \cite{zhang2022rethinking}, Smooth Layers
        & $34.97\%$ & \cite{shi2021fast}, Interval
        & $58.4\%$ & \cite{globally2021leino}, General Lipschitz
        & $71\%^*$ & \cite{Zhai2020MACER:}, Nym.-Prsn. 
        & $18\%^*$ & \cite{lecuyer2019certified}, Differential Privacy
        & $5\%^*$ & \cite{lecuyer2019certified}, Differential Privacy \\
        \hline
        
        \multirow{6}{*}{ImageNet} & \multicolumn{2}{c|}{$\epsilon=1/255$} & \multicolumn{2}{c|}{\multirow{6}{*}{\shortstack{No work achieves\\ $>0\%$ certified accuracy\\ under large $\epsilon$ yet.}}} & \multicolumn{2}{c|}{$\epsilon=1.0$} & \multicolumn{2}{c|}{$\epsilon=2.0$} & \multicolumn{2}{c|}{$\epsilon=1.0$} & \multicolumn{2}{c}{$\epsilon=2.0$} \\
        \cline{2-3} \cline{6-13}
        
        & $\mathbf{38.2\%}^*$ & \cite{salman2019provably}, Nym.-Prsn. & & 
        & $\mathbf{67.0\%}^*$ & \cite{xiao2023densepure}, Nym.-Prsn.
        & $\mathbf{42.2\%}^*$ & \cite{xiao2023densepure}, Nym.-Prsn.
        & $\mathbf{55\%}^*$ & \cite{yang2020randomized}, Nym.-Prsn. 
        & $\mathbf{48\%}^*$ & \cite{yang2020randomized}, Nym.-Prsn. \\
        
        & $28.6\%^*$ & \cite{cohen2019certified}, Nym.-Prsn. & \multicolumn{2}{c|}{} 
        & $54.3\%^*$ & \cite{carlini2023certified}, Nym.-Prsn.
        & $30.4\%^*$ & \cite{yang2021certified}, Nym.-Prsn.
        & $49\%$ & \cite{levine2021improved}, Lipschitz
        & $45\%$ & \cite{levine2021improved}, Lipschitz \\
        
        & \multicolumn{2}{c|}{} & & 
        & $45\%^*$ & \cite{salman2019provably}, Nym.-Prsn.
        & $29.5\%^*$ & \cite{carlini2023certified}, Nym.-Prsn.
        & $42\%^*$ & \cite{Zhang2020BlackBoxCW}, Nym.-Prsn.
        & $30\%^*$ & \cite{Zhang2020BlackBoxCW}, Nym.-Prsn.\\
        
        & \multicolumn{2}{c|}{} & & 
        & $44.6\%^*$ & \cite{horvath2022boosting}, Nym.-Prsn.
        & $28.6\%^*$ & \cite{horvath2022boosting}, Nym.-Prsn. 
        & $40\%^*$ & \cite{teng2020ell}, Nym.-Prsn.
        & $26\%^*$ & \cite{teng2020ell}, Nym.-Prsn. \\
        
        & \multicolumn{2}{c|}{} & & 
        & $44.4\%^*$ & \cite{yang2021certified}, Nym.-Prsn. 
        & $28\%^*$ & \cite{salman2019provably}, Nym.-Prsn. 
        & $25\%^*$ & \cite{lecuyer2019certified}, Differential Privacy
        & $16\%^*$ & \cite{lecuyer2019certified}, Differential Privacy \\
        
        \bottomrule 
    \end{tabular}
    }
    \label{tab:leaderboard}
    \vspace{-1.5em}
\end{table*}

%% file: contents/appendix.tex
\appendices



\section{Scalability and Tightness Measurements}
    \label{adxsec:scalability-tightness}
    
    This appendix contains more discussion on the scalability and tightness characterization in \Cref{sec:main-taxonomy}.
    
    \para{Scalability measured by time complexity.}
    Note that the time complexity for DNN inference is $O(lw^2)$.
    As we can see in \Cref{tab:property-table}, all complete verification approaches have exponential time complexity $O(2^{lw})$ which coincides with the theoretical scalability barriers~\cite{katz2017reluplex,weng2018towards}.
    The $\poly(l,w)$ means a time complexity higher than $O(lw^3)$.
    All approaches for smoothed DNNs have complexity $O(Slw^2)$, which is because the sampling time cost is much higher than the actual bound computation whose time complexity is subsumed.

    \para{Deails on tightness ranks.}
    For general DNNs, we rank the tightness from $T_1$ to $T_7$ where $T_7$ is the tightest. 
    $T_1 < T_2 < T_3$ comes from benchmark results, $T_3 < T_4 < T_5$ comes from theoretical analyses~\cite{salman2019convex}, and $T_5 < T_6$ and $T_6 < T_7$ come from empirical observations in \cite{dathathri2020enabling} and \cite{muller2022prima} respectively.
    For smoothed DNNs we rank the tightness from $ST_1$ to $ST_4$ based on existing theoretical analyses: $ST_1 < ST_2$ comes from \cite{Dvijotham2020A,li2019certified}, $ST_2 < ST_3$ comes from  \cite{Dvijotham2020A,salman2019provably,yang2020randomized}, and
    $ST_3 < ST_4$ comes from \cite{mohapatra2020higher}.

\section{Omitted Illustrations}
    \label{adxsec:omitted-illustration}
    This appendix includes the omitted figure illustrations.
    
    \subsection{Perturbation Region of \texorpdfstring{$\ell_p$}{Lp} Adversary}
        \label{adxsubsec:lp-adversary-fig}
        
        \begin{figure}[!h]
            \centering
            \includegraphics[width=0.35\textwidth]{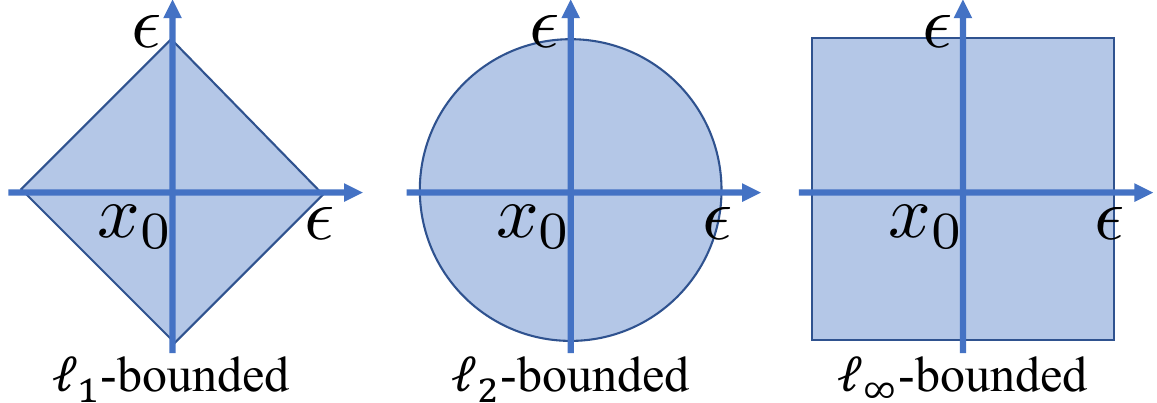}
            \vspace{-0.5em}
            \caption{An $(\ell_p,\epsilon)$-adversary crafts perturbed input from $\cL_p$-bounded region centered at clean input $x_0$. From left to right are $\ell_1$-, $\ell_2$-, and $\ell_\infty$-bounded perturbation regions in 2D space with radius $\epsilon$.}
            \label{fig:different-norms}
            \vspace{-1.0em}
        \end{figure}

    \subsection{ReLU Relaxation with Single Input Variable}
        \label{adxsubsec:single-illustration}
        
        \begin{figure}[H]
            \centering
            \vspace{-1.0em}
            \begin{subfigure}{.2\linewidth}
                 \centering
                 \includegraphics[width=\linewidth]{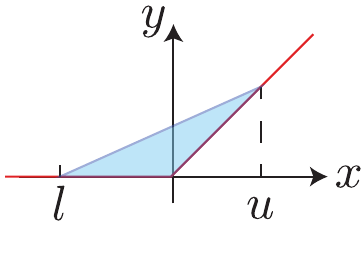}
                 \vspace{-1.45em}
                 \caption{\,}
                 \label{fig:tightest-linear-polytope}
             \end{subfigure}
            \begin{subfigure}{.21\linewidth}
                 \centering
                 \includegraphics[width=\linewidth]{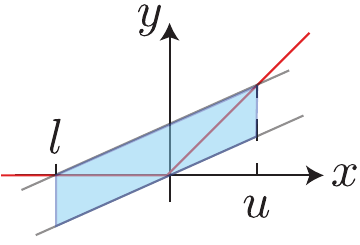}
                 \vspace{-1.45em}
                 \caption{\,}
                 \label{fig:relu-lower-bound-1}
             \end{subfigure}
             \begin{subfigure}{.2\linewidth}
                 \centering
                 \includegraphics[width=\linewidth]{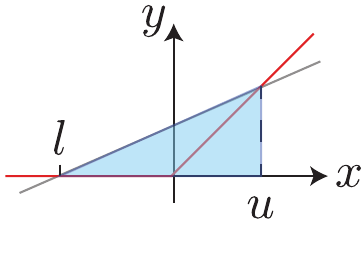}
                 \vspace{-1.45em}
                 \caption{\,}
                 \label{fig:relu-lower-bound-2}
             \end{subfigure}
             \begin{subfigure}{.23\linewidth}
                 \centering
                 \includegraphics[width=\linewidth]{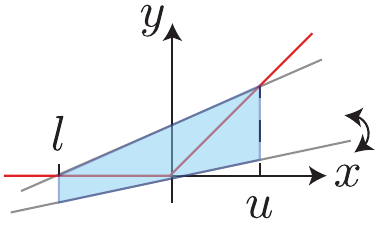}
                 \vspace{-1.45em}
                 \caption{\,}
                 \label{fig:relu-lower-bound-3}
             \end{subfigure}
            \caption{Different linear relaxations for ReLU shown as blue region. 
            (a) shows the tightest single-neuron polytope for ReLU which is used in \cite{ehlers2017formal,wong2018provable,wong2018scaling,tjeng2018evaluating};
            (b) is used in \cite{singh2019abstract,weng2018towards,zhang2018efficient};
            (c) is used in \cite{singh2019abstract,zhang2018efficient}; 
            (d) shows that the slope of lower bound can be dynamically adjusted or optimized~\cite{lyu2020fastened,singh2019abstract,xu2021fast,zhang2018efficient}. }
            \label{fig:different-relu-lower-bounds}
        \end{figure}
    
    \subsection{ReLU Relaxation with Multiple Input Variables}
        \label{adxsec:multi-neuron-illustration}
    
        \begin{figure}[!h]
            \vspace{-1em}
            \centering
            \begin{subfigure}{.47\linewidth}
                 \centering
                 \includegraphics[width=0.5\linewidth]{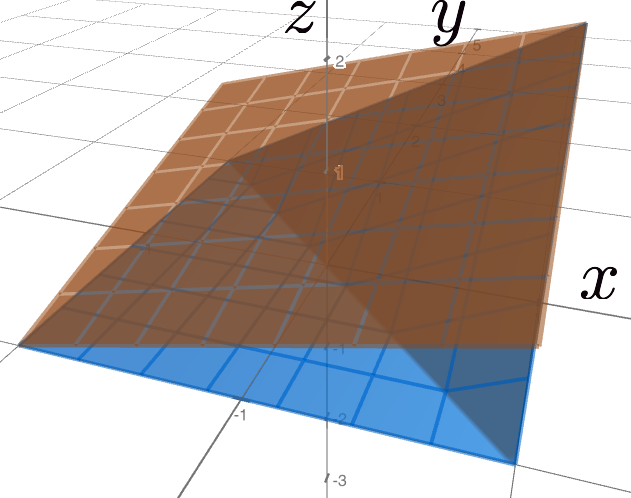}
                 \vspace{-0.75em}
                 \caption{Single ReLU polytope (shown in brown) gives looser upper bound for $z$.}
                 \label{fig:multi-neuron-1}
             \end{subfigure}
            \begin{subfigure}{.49\linewidth}
                 \centering
                 \includegraphics[width=0.5\linewidth]{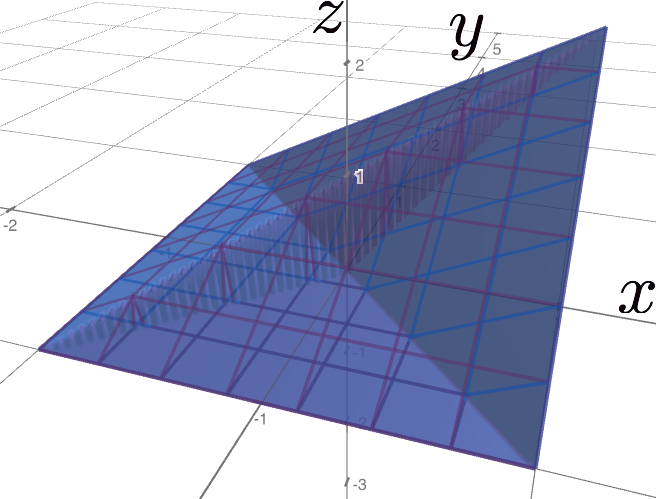}
                 \vspace{-0.75em}
                 \caption{Multi-neuron convex relaxation gives tighter upper bound for $z$ (shown as two dark blue facets).}
                 \label{fig:multi-neuron-2}
             \end{subfigure}
             \vspace{-0.5em}
             \caption{Comparison of convex relaxation for $z = \relu(x+y)$~(shown as bottom blue surface in (a)), where $x,y\in [-1,1]$. Vertical axis is the $z$-axis.}
            \label{fig:multi-neuron}
            \vspace{-1.3em}
        \end{figure}
\section{Details on Linear Inequality Based Verification}
    \label{adxsec:polyhedra-illustration}
    This appendix entails the omitted details of linear inequality verification approaches introduced in \Cref{sec:linear-inequality-propagation-intro}.

    \para{More details on polyhedra abstraction.}
        \t{Fast-Lin}~\cite{weng2018towards} uses a parallel line as the lower bound as shown in \Cref{fig:relu-lower-bound-1}.
        \t{CROWN}~\cite{zhang2018efficient} and \t{DeepPoly}~\cite{singh2019abstract} both support adjustable lower bound.
        They both use $y = \lambda x$ with adjustable $\lambda \in [0,1]$ as the lower bound, while their heuristics for determining $\lambda$ are slightly different.
        \t{FROWN}~\cite{lyu2020fastened} and \t{$\alpha$-CROWN}~\cite{xu2021fast} deploy gradient-based optimization on lower bound slope $\lambda$ to improve tightness.
        
        These approaches maintain the linear bound for each layer~$k$ in the form of $\bL_k x + b_{L,k} \le z_k(x) \le \bU_k x + b_{U,k}$ for any $x \in B_{p,\epsilon}(x_0)$.
        From the bound for layer $k$, we can deduct the bound after affine mapping $\hz_{k+1} = \bW_k z_k + b_k$:
        \begin{equation}
            \small
            \begin{aligned}
                & (\bW_k^+ \bL_k + \bW_k^- \bU_k)x + \bW_k^+ b_{L,k} + \bW_k^- b_{U,k} + b_k \\
                \le & \hz_{k+1}(x) \\
                \le & (\bW_k^+ \bU_k + \bW_k^- \bL_k)x + \bW_k^+ b_{U,k} + \bW_k^- b_{L,k} + b_k.
            \end{aligned}
            \label{eq:affine-linear-bound}
        \end{equation}
        Then, they compute the activation value bound $l_{k+1}$ and $u_{k+1}$ for $\hz_{k+1}$, and compute the linear bound for $z_{k+1}(x) = \relu(\hz_{k+1}(x))$ using ReLU lower and upper bound respectively.
        By repeating the process, they finally bound the last layer $\hz_l$, i.e., the model $f$ itself.

    \para{Zonotope abstraction.}
    Zonotope is another type of over-approximation or abstract interpretation domain that can be propagated layer by layer efficiently~\cite{gehr2018ai2,mirman2018differentiable,singh2018fast,anderson2019optimization,singh2018robustness}.
    Zonotope abstraction has the same efficiency and slightly inferior tightness compared to polyhedra abstraction~\cite{salman2019convex}.

    \para{Duality-based approaches.}
    \label{sec:dual-intro}
    Since the robustness verification can be viewed as an optimization problem~(\Cref{prob:verification-optimization-problem}), we can consider its Lagrangian dual problem.
    Especially, since \Cref{prob:verification-optimization-problem} is a minimization problem, any feasible dual solution provides a valid lower bound of the primal problem and therefore a valid verification.
    Moreover, the dual problem is always convex~\cite{boyd2004convex}.
    %
    Typical duality-based approaches are \t{WK}~\cite{wong2018provable,wong2018scaling}, \t{D-LP}~\cite{dvijotham2018dual}, \t{PVT}~\cite{dvijotham2018training}, and \t{Lagrangian decomposition}~\cite{bunel2020lagrangian} where \t{WK} is proved to share equivalent tightness with polyhedra abstraction approaches, and the others are proved to share equivalent tightness with linear programming based approaches~\cite{salman2019convex}.

\section{Illustration of Robust Training Approaches}
    \label{adxsec:robust-training-approaches}

    \para{Regularization-based training.}
    For complete verification, Xiao~et~al~\cite{xiao2018training} find that the number of branches is upper bounded by the number of unstable neurons~(see \Cref{def:active-inactive-relu-neuron}) which motivates a regularization term to increase the ReLU neuron's stability for training.
    For complete verification based on linear region traversal, we can train with a regularization term maximizing the margin to non-robust regions~\cite{croce2019provable,Croce2020Provable}.
    The Lipschitz and curvature verification favor small Lipschitz constant and small curvature bounds respectively.
    Therefore, the corresponding robust training approaches explicitly penalize large Lipschitz or curvature bounds~\cite{lee2020lipschitz,globally2021leino,singla2020second,tsuzuku2018lipschitz}.
    
    \para{Relaxation-based training.}
    For linear relaxation based verification approaches, models with tight linear relaxation bounds are favored.
    To train such models, corresponding robust training approaches usually use the computed bounds from linear relaxation as the training objective to explicitly improve the bound tightness. 
    This idea is similar to the powerful empirical defense named adversarial training~\cite{madry2017towards} which uses effective attacks to approximately find ``most adversarial'' example $\max_{x\in B_{p,\epsilon}(x_0)} \gL(f_\theta(x), y_0)$ and minimize model weights $\theta$ w.r.t. it.
    In relaxation-based training, instead, we compute an upper bound of $\max_{x\in B_{p,\epsilon}(x_0)} \gL(f_\theta(x), y_0)$ and minimize it.
    The bound can be derived from IBP~\cite{gowal2019scalable,shi2021fast}, polyhedra-based~\cite{Balunovic2020Adversarial,lyu2020fastened,zhang2020towards}, zonotope-based~\cite{mirman2018differentiable}, or duality-based verification~\cite{li2019robustra,wong2018provable,dvijotham2018training}.
    Some useful training tricks are: combining relaxation-based loss with standard loss to improve benign accuracy~\cite{gowal2019scalable,wang2018mixtrain,zhang2020towards}, applying relaxation on some layers but not all to balance benign accuracy and certified robustness~\cite{Balunovic2020Adversarial}, specialized weight initialization and training scheduling~\cite{shi2021fast}, and using reference space to guide the relaxation~\cite{li2019robustra}.
    An intriguing phenomenon of relaxation-based training is that tighter relaxation, when used as the training objective, may not lead to more certifiably robust models~\cite{jovanovic2021certified}, while the loosest IBP relaxation can achieve almost the highest certified robustness.
    A conjecture is that tighter relaxation may lead to a less smooth loss landscape containing discontinuities or sensitive regions which poses challenges for gradient-based training~\cite{jovanovic2021certified,lee2021towards}.
    Theoretical understanding of relaxation-based training is still lacking.
    Note that solver based and branch-and-bound based complete verification usually use linear relaxations for bounding.
    Therefore, models trained with these relaxation-based training approaches can usually be efficiently certified by these complete verification approaches~\cite{tjeng2018evaluating,wang2021beta}.
    
    \para{Augmentation-based training.}
    Since randomized smoothing based verification favors models to perform well for noisy inputs, to obtain high certified robustness, we can train the DNNs with noisy inputs, resulting in augmentation-based training~\cite{cohen2019certified,lecuyer2019certified,li2019certified}.
    Built upon such augmentation-based training, later approaches combine augmentation with regularization terms to encourage the prediction stability/consistency when the input noise is added~\cite{jeong2020consistency,jeong2021smoothmix,Zhai2020MACER:}.
    Strategic training regularization combined with augmentation and ensemble is effective and achieves the state-of-the-art certified robustness against $\ell_2$ adversary~\cite{horvath2022boosting,yang2021certified}.
    Adversarial training combined with augmentation~\cite{salman2019provably}, and training unlabeled data~\cite{carmon2019unlabeled} are also shown effective.
    Recently, diffusion models~\cite{song2021scorebased}, which intrinsically possess the denoising ability, are leveraged to build models for randomized smoothing~\cite{carlini2023certified,xiao2023densepure}.
    They achieve superior or competitive certified robustness compared to above methods though require large model size which results in large inference overhead.

\section{Benchmark Evaluation Details}
    \label{adxsec:benchmark-eval-detail}

    \input{tables/final_main_table}

    \input{tables/final-main-table-2}

    \para{Experiment environment.} Our toolkit implementation is based on \t{PyTorch}~\cite{pytorch}.
    In the toolkit, we tend to integrate the original implementations released by the authors when it is available;
    otherwise, we implement and optimize them to match the reported performance.
    We run the evaluation on a $24$-core Intel Xeon Platinum 8259CL CPU running at \SI{2.50}{GHz} with a single NVIDIA Tesla T4 GPU.
    
    \subsection{Comparison of Deterministic Verification} 
        \label{sec:exp-A}

        We present a thorough comparison of representative \emph{deterministic verification approaches} in \Cref{tab:final_main_table}.
        
        We evaluate on $7$ different DNNs on CIFAR-10.
        Among them, $3$ models~(\sc{FCNNa} - \sc{FCNNc}) are fully-connected networks, and $4$ models~(\sc{CNNa} - \sc{CNNd}) are convolutional neural networks.
        The number of neurons ranges from $50$~(\sc{FCNNa}) to about $200,000$~(\sc{CNNd}).
        For each DNN structure, we train two sets of weights: \texttt{adv}---PGD adversarial training with $\epsilon = 8/255$; \texttt{cadv}---\t{CROWN-IBP} training with $\epsilon = 8/255$, where $\epsilon$ is the $\cL_\infty$ attack radius.
        The PGD adversarial training~\cite{madry2017towards} is a strong empirical defense, and \t{CROWN-IBP}~\cite{zhang2020towards} is a strong robust training approach.
        For PGD adversarial training, following the literature~\cite{madry2017towards,tramer2020adaptive}, we set the attack step size to be $\epsilon / 50$, attack iterations to be $100$ with random initialization, and train for $40$ epochs with $0.1$ learning rate and SGD optimizer.
        For \t{CROWN-IBP}, we use the official code release~\cite{zhang2020towards} and default hyperparameters: $100$ epochs with Adam optimizer and $5\times 10^{-4}$ learning rate on MNIST, and $200$ epochs with SGD optimizer and $0.001$ learning rate on CIFAR-10.
        More hyperparameters can be found in our open-source toolbox.
        We choose these training configurations to reflect two common types of models on which verification approaches are used: empirically defended models and robustly trained models.
        All models are trained to reach their expected robustness as reported in the corresponding papers.
        We defer the detailed model structure and statistics to our website.

\begin{table*}[ht]
    \centering
    \caption{\small $\ell_\infty$ certified robust accuracy w.r.t. different radii $r$'s~(our method shown in gray). 
    }
    \resizebox{\linewidth}{!}{
    \begin{tabular}{c|c|c|c|cccccccccccc}
        \toprule
        \multirow{2}{*}{Dataset} & \multirow{2}{*}{Model} & Certification & Clean & \multicolumn{12}{c}{Certified Accuracy under Radius $r$} \\
        \cline{5-16}
        & & Approach & Accuracy &   1/255   &   2/255   &   3/255   &   4/255   &   5/255   &   6/255   &   7/255   &    8/255   &   9/255   &   10/255   &   11/255   &   12/255   \\
        \hline\hline
        \multirow{4}{*}{MNIST} & Gaussian  & Neyman-Pearson & 
        \multirow{2}{*}{  99.1\%  } &	  \bf 98.1\%   &	  97.4\%   &	  \bf 96.6\%   &	  95.8\%   &	  \bf 95.2\%   &	  92.4\%   &	  89.4\%   &	  85.2\%   &	  80.8\%   &	  73.2\%   &	  64.0\%   &	  50.7\%  	 \\
        \cline{3-3} \cline{5-16}
        & Augmentation & \cellcolor{tabgray}\textbf{Our Method} &
         &	\cellgray   \bf 98.1\%   &	\cellgray   \bf 97.5\%   &	\cellgray   \bf 96.6\%   &	\cellgray   \bf 96.1\%   &	\cellgray   \bf 95.2\%   &	\cellgray   \bf 92.7\%   &	\cellgray   \bf 90.5\%   &	\cellgray   \bf 86.8\%   &	\cellgray   \bf 82.8\%   &	\cellgray   \bf 77.6\%   &	\cellgray   \bf 68.8\%   &	\cellgray   \bf 60.0\%  	 \\	
        \cline{2-16}
        & Consistency & Neyman-Pearson & 
        \multirow{2}{*}{  98.5\%  } &	  \bf 98.3\%   &	  \bf 98.2\%   &	  \bf 97.2\%   &	  \bf 96.4\%   &	  95.4\%   &	  93.9\%   &	  91.5\%   &	  88.3\%   &	  83.9\%   &	  78.7\%   &	  71.2\%   &	  62.7\%   \\
        \cline{3-3} \cline{5-16}
        & \cite{jeong2020consistency} & \cellcolor{tabgray}\textbf{Our Method} & 
         &	\cellgray   \bf 98.3\%   &	\cellgray   \bf 98.2\%   &	\cellgray   \bf 97.2\%   &	\cellgray   \bf 96.4\%   &	\cellgray   \bf 95.6\%   &	\cellgray   \bf 94.3\%   &	\cellgray   \bf 92.7\%   &	\cellgray   \bf 89.0\%   &	\cellgray   \bf 86.0\%   &	\cellgray   \bf 81.9\%   &	\cellgray   \bf 75.5\%   &	\cellgray   \bf 67.5\%  	 \\
        \hline\hline
        & & & Clean & \multicolumn{12}{c}{Certified Accuracy under Radius $r$} \\
        \cline{5-16}
        & & & Accuracy &  0.5/255  &  1/255  &  1.5/255  &  2/255  &  2.5/255  &  3/255  &  3.5/255  &  4/255  &  4.5/255  &  5/255  &  5.5/255  &  6/255  \\
        \hline
        \multirow{4}{*}{CIFAR-10} & Gaussian  & Neyman-Pearson & 
        \multirow{2}{*}{ 65.6\% } &	 52.0\%  &	 45.3\%  &	 41.1\%  &	 36.3\%  &	 32.6\%  &	 26.7\%  &	 21.9\%  &	 18.1\%  &	 15.1\%  &	 10.9\%  &	 8.9\%  &	 6.1\% 	  \\
        \cline{3-3} \cline{5-16}
        & Augmentation & \cellcolor{tabgray}\textbf{Our Method} &
         &	\cellgray  \bf 52.3\%  &	\cellgray  \bf 45.6\%  &	\cellgray  \bf 41.5\%  &	\cellgray  \bf 37.6\%  &	\cellgray  \bf 33.8\%  &	\cellgray  \bf 28.8\%  &	\cellgray  \bf 23.7\%  &	\cellgray  \bf 19.5\%  &	\cellgray  \bf 17.2\%  &	\cellgray  \bf 13.9\%  &	\cellgray  \bf 10.5\%  &	\cellgray  \bf 8.1\% 	\\
        \cline{2-16}
        & Consistency & Neyman-Pearson & 
        \multirow{2}{*}{ 52.6\% } &	 47.1\%  &	 \bf 45.5\%  &	 \bf 43.6\%  &	 40.6\%  &	 38.3\%  &	 36.0\%  &	 33.4\%  &	 30.5\%  &	 28.5\%  &	 25.2\%  &	 22.0\%  &	 20.3\% 	\\
        \cline{3-3} \cline{5-16}
        & \cite{jeong2020consistency} & \cellcolor{tabgray}\textbf{Our Method} & 
         &	\cellgray  \bf 47.2\%  &	\cellgray  \bf 45.5\%  &	\cellgray  \bf 43.6\%  &	\cellgray  \bf 40.9\%  &	\cellgray  \bf 38.9\%  &	\cellgray  \bf 36.9\%  &	\cellgray  \bf 34.5\%  &	\cellgray  \bf 31.9\%  &	\cellgray  \bf 29.5\%  &	\cellgray  \bf 28.1\%  &	\cellgray  \bf 24.9\%  &	\cellgray \bf 22.0\% \\
        \bottomrule
    \end{tabular}
    }
    \label{tab:ell-infty}
    \vspace{-1em}
\end{table*}
        \para{Evaluation protocol.}
        We measure the performance of verification approaches by their  \textbf{certified accuracy} w.r.t. $\cL_\infty$ radius $\epsilon=8/255$.
        $\cL_\infty$ adversary is supported by most deterministic verification approaches.
        The certified accuracy, as a measurement of certified robustness, is defined as 
        \begin{equation}
            \small
            \mathrm{certacc} := \dfrac{\text{\# samples verified to be robust}}{\text{\# number of all samples}}.
            \label{eq:certacc}
        \end{equation}
        On each dataset, we uniformly sample $100$ test samples as the fixed set for evaluation.
        We limit the running time to \SI{60}{s} per instance~(so that verifying all $14$ benchmark models with each approach takes about one day) and count timeout instances as ``not verified'' to favor efficient and practical verification approaches.
         This time limit is aligned with common settings. 
         For example, the recent competition~(VNN-COMP 2021~\cite{ashtiani2020black}) for complete verification sets 6-hour as the time limit.
         For a fair comparison, we relax this time limit from 6 hours to one day since we benchmark multiple models together.
        Moreover, running tools with the one-day time limit per approach takes overall around 2.5 months considering around 20 approaches and all settings. Therefore, for time and energy concerns we did not benchmark with longer time limits. Practical users can explore other time limits with our open-source toolkit.
        We also report the robust accuracy under empirical attack~(PGD attack with $100$ steps, step size $\epsilon/50$, and random starts following \cite{madry2017towards,tramer2020adaptive}), which upper bounds certified accuracy.

            %
        
            %
        \Cref{tab:final_main_table} shows certified accuracy on CIFAR-10 for deterministic approaches.
        Each row corresponds to a verification approach, PGD attack, or clean accuracy.
        More results such as average certified robustness radius, average running time, and results on MNIST are on our website.
        Findings from our evaluation are discussed in \Cref{subsec:benchmark-evaluation}.

    \subsection{Comparison of Probabilistic Verification}
    
    
        
        We present a thorough comparison of representative \emph{probabilistic verification approaches} for smoothed DNNs with different smoothing distributions and robust training approaches.
        %
        We either fix the robust training part and vary the verification approaches or the other way around.
        
        \para{Evaluation protocol.}
        We use ResNet-110 and Wide ResNet 40-2 as the model architecture.
        $n=1,000$ samples are used for selecting the top label; $N = 100,000$ samples are used for certification.
        For all robust training approaches, we adopt default hyperparameters as reported in corresponding papers.
        The failure probability is set to $1 - \alpha = .001$. 
        We uniformly draw $500$ samples from the test set for evaluation.
        All the above settings follow common practice in \cite{cohen2019certified,yang2020randomized}.
        
        \para{Comparison results and discussion.}
        We show results on CIFAR-10 in \Cref{tab:exp-B-main-table}.
        Results on ImageNet can be found on our benchmark website.
        Findings from our evaluation are discussed in \Cref{subsec:benchmark-evaluation}.




\section{Tighter Certification against \texorpdfstring{$\ell_\infty$}{Linf} Adversary}

\label{adxsec:double-sampling-l-inf}
We extend the very recent double sampling randomized smoothing in \cite{li2022double} to provide robustness certification for smoothed DNNs by sampling the statistics of the smoothed DNNs' prediction using both the original smoothing distribution $\gP$ and an additional smoothing distribution $\gQ$ that shares the same form but a different variance from $\gP$'s variance.
Note that we leverage additional information---the prediction probability under $\gQ$.
In contrast, the zeroth-order methods only leverage the sampling probability information from $\gP$.
The extension methodology is listed in Appendix~H.3 of \cite{li2022double}.

    Now we systematically evaluate our extension of the double sampling method and demonstrate that it achieves tighter certification than the classical Neyman-Pearson-based certification~(the tightest zeroth-order information approach) against $\ell_\infty$-bounded perturbations on MNIST and CIFAR-10.
    
        
        \noindent\textbf{Smoothing Distributions.}
        For a given distribution $\gD$, we let $\Std(\gD)$ be its average component-wise standard deviation: $\Std(\gD) := \sqrt{\frac{1}{d} \E_{\vdelta\sim\gD} [\|\vdelta\|_2^2]}$~as first used in \cite{yang2020randomized}.
        We set $\Std(\gP) = 0.75$, $\Std(\gQ) = 0.6$ on both MNIST and CIFAR-10.
        We use generalized Gaussian as the smoothing distribution following \cite{li2022double} where $d - k = 8$ on MNIST and $d - k = 12$ on CIFAR-10.
        Note that we did not finetune these hyperparameters and we expect the existence of better hyperparameters.
        
        \noindent\textbf{Models.}
        We train the models using both commonly-used Gaussian augmentation~\cite{cohen2019certified} and state-of-the-art Consistency training~\cite{jeong2020consistency}. 
        On all datasets, we use the default model structures and hyperparameters.
        All models are trained with the original smoothing distribution $\gP$.
        
        \noindent\textbf{Baselines.}
        We consider the Neyman-Pearson-based certification method as the baseline.
        For both baseline and our method,
        we set the certification confidence to be $1 - 2\alpha = 99.8\%$.
        We use $10^5$ samples for estimating $P_A$ and $Q_A$ per instance.
        Note that Neyman-Pearson certification does not use the information from additional distribution and all $10^5$ samples are used to estimate the interval of $P_A$.
        In our method, we use $5\times 10^4$ samples to estimate the interval of $P_A$ and the rest $5\times 10^4$ samples for $Q_A$.
        
        \noindent\textbf{Metric.}
        We uniformly draw $1000$ samples from the test set, and report the \emph{certified accuracy} under each radius $r$ as defined in \Cref{eq:certacc}.
        We also report the benign accuracy of the smoothed classifier.
        Both settings and the metric follow the standard evaluation protocol in  literature~\cite{cohen2019certified,yang2020randomized}.

    \noindent\textbf{Main Results.}
        The experimental results for certification against $\ell_\infty$ adversary are shown in \Cref{tab:ell-infty}.
        %
        We observe that, for \emph{all} evaluated models, our method yields significantly higher certified accuracy.
        For example, when $r = 12/255$ our method improves the MNIST robust accuracy from $50.7\%$ to $60.0\%$;
        when $r = 5/255$ our method improves CIFAR-10 robust accuracy from $25.2\%$ to $28.1\%$.
        Thus, leveraging additional information can indeed provide tighter robustness certification over zeroth-order certification approaches for smoothed DNNs not only against $\ell_1$ and $\ell_2$ adversaries but also against $\ell_\infty$ adversary.

%% file: tables/final_main_table.tex

\begin{table*}[ht]
    \caption{\emph{Certified accuracy} on CIFAR-10  certified by  \emph{deterministic verification approaches} for DNNs. 
    Failing to verify or exceeding \SI{60}{s} verification time limit is counted as ``non-robust''.
    $0\%$ certified accuracy means too loose or too slow to verify all input samples.
    The verification is against $\cL_\infty$ adversary with radius $\epsilon = 8/255$. We include model accuracy under PGD attack as the upper bound of the certified accuracy. 
    The \textbf{bolded} numbers mark the highest ones among verification approaches.}
    \centering
    \vspace{-1.0em}
        \newcommand{\mytabwidth}{0.9}
    \resizebox{\mytabwidth\textwidth}{!}{
    \begin{tabular}{c|c|c|c|c|c|c|c|c|c|c|c|c|c|c|c|c|c|c}
\toprule
\multicolumn{5}{c|}{Verification Approach}   &         \mc{2}{\sc{FCNNa}} &         \mc{2}{\sc{FCNNb}} &         \mc{2}{\sc{FCNNc}} &          \mc{2}{\sc{CNNa}} &          \mc{2}{\sc{CNNb}} &          \mc{2}{\sc{CNNc}} &         \emc{2}{\sc{CNNd}}\\
\hline
\multicolumn{4}{c|}{Category} & Name & \texttt{adv} & \texttt{cadv} & \texttt{adv} & \texttt{cadv} & \texttt{adv} & \texttt{cadv} & \texttt{adv} & \texttt{cadv} & \texttt{adv} & \texttt{cadv} & \texttt{adv} & \texttt{cadv} & \texttt{adv} & \texttt{cadv}\\
\hline\hline
\multirow{2}{*}{Complete} & \multicolumn{3}{c|}{Solver-Based} & \t{Bounded MILP}~\cite{tjeng2018evaluating} &   $\mb{19\%}$ &   $\mb{27\%}$ &         $1\%$ &   $\mb{25\%}$ &   $\tle{0\%}$ &   $\tle{0\%}$ &   $\tle{0\%}$ &   $\mb{34\%}$ &   $\tle{0\%}$ &   $\mb{36\%}$ &   $\tle{0\%}$ &   $\tle{0\%}$ &   $\tle{0\%}$ &   $\tle{0\%}$ \\
 \cline{2-19}
& \multicolumn{3}{c|}{Branch-and-Bound} &  \t{AI$^2$}~\cite{gehr2018ai2} &   $\mb{19\%}$ &   $\mb{27\%}$ &    $\mb{7\%}$ &        $23\%$ &   $\tle{0\%}$ &        $22\%$ &    $\mb{8\%}$ &   $\mb{34\%}$ &   $\tle{0\%}$ &        $20\%$ &   $\tle{0\%}$ &        $14\%$ &   $\tle{0\%}$ &   $\tle{0\%}$ \\
\hline
\multirow{15}{*}{Incomplete} & \multirow{10}{*}{\shortstack{Linear\\ Relaxtion}} & \multicolumn{2}{c|}{Linear Programming} &  \t{LP-Full}~\cite{salman2019convex,weng2018towards} &        $15\%$ &   $\mb{27\%}$ &         $6\%$ &   $\mb{25\%}$ &   $\tle{0\%}$ &   $\tle{0\%}$ &   $\tle{0\%}$ &   $\tle{0\%}$ &   $\tle{0\%}$ &   $\tle{0\%}$ &   $\tle{0\%}$ &   $\tle{0\%}$ &   $\tle{0\%}$ &   $\tle{0\%}$ \\
 \cline{3-19}
& & \multirow{8}{*}{\shortstack{Linear\\ Inequality}} & Interval &  \t{IBP}~\cite{gowal2019scalable} &   $\tle{0\%}$ &   $\mb{27\%}$ &   $\tle{0\%}$ &   $\mb{25\%}$ &   $\tle{0\%}$ &   $\mb{30\%}$ &   $\tle{0\%}$ &   $\mb{34\%}$ &   $\tle{0\%}$ &        $35\%$ &   $\tle{0\%}$ &   $\mb{38\%}$ &   $\tle{0\%}$ &   $\mb{28\%}$ \\
 \cline{4-19}
& & & \multirow{6}{*}{Polyhedra} &  \t{Fast-Lin}~\cite{weng2018towards} &        $15\%$ &        $25\%$ &         $4\%$ &        $18\%$ &   $\tle{0\%}$ &        $19\%$ &         $3\%$ &        $26\%$ &   $\tle{0\%}$ &        $15\%$ &   $\tle{0\%}$ &         $7\%$ &   $\tle{0\%}$ &   $\tle{0\%}$ \\
\cline{5-19}
 
& & & &  \t{CROWN}~\cite{zhang2018efficient} &        $15\%$ &   $\mb{27\%}$ &         $6\%$ &        $20\%$ &   $\tle{0\%}$ &        $22\%$ &    $\mb{8\%}$ &        $33\%$ &    $\mb{1\%}$ &        $20\%$ &   $\tle{0\%}$ &   $\tle{0\%}$ &   $\tle{0\%}$ &   $\tle{0\%}$ \\
\cline{5-19}
& & & &  \t{CNN-Cert}~\cite{boopathy2019cnn} &        $15\%$ &   $\mb{27\%}$ &         $5\%$ &        $20\%$ &   $\tle{0\%}$ &   $\tle{0\%}$ &         $7\%$ &        $33\%$ &   $\tle{0\%}$ &        $20\%$ &   $\tle{0\%}$ &   $\tle{0\%}$ &   $\tle{0\%}$ &   $\tle{0\%}$ \\
\cline{5-19}
 
& & & & \t{CROWN-IBP}~\cite{zhang2020towards} &         $9\%$ &   $\mb{27\%}$ &   $\tle{0\%}$ &        $22\%$ &   $\tle{0\%}$ &        $28\%$ &   $\tle{0\%}$ &   $\mb{34\%}$ &   $\tle{0\%}$ &        $31\%$ &   $\tle{0\%}$ &        $32\%$ &   $\tle{0\%}$ &        $25\%$ \\
\cline{5-19}
& & & &  \t{DeepPoly}~\cite{singh2019abstract} &        $15\%$ &   $\mb{27\%}$ &         $6\%$ &        $20\%$ &   $\tle{0\%}$ &        $22\%$ &    $\mb{8\%}$ &        $33\%$ &    $\mb{1\%}$ &        $20\%$ &   $\tle{0\%}$ &         $7\%$ &   $\tle{0\%}$ &   $\tle{0\%}$ \\
\cline{5-19}

& & & & \t{RefineZono}~\cite{singh2018robustness} &   $\tle{0\%}$ &   $\mb{27\%}$ &   $\tle{0\%}$ &   $\tle{0\%}$ &   $\tle{0\%}$ &   $\tle{0\%}$ &   $\tle{0\%}$ &   $\tle{0\%}$ &   $\tle{0\%}$ &   $\tle{0\%}$ &   $\tle{0\%}$ &   $\tle{0\%}$ &   $\tle{0\%}$ &   $\tle{0\%}$ \\
\cline{4-19}
 
& & & Duality &  \t{WK}~\cite{wong2018provable,wong2018scaling}  &        $15\%$ &        $25\%$ &         $4\%$ &        $18\%$ &   $\tle{0\%}$ &        $19\%$ &         $3\%$ &        $26\%$ &   $\tle{0\%}$ &        $15\%$ &   $\tle{0\%}$ &         $7\%$ &   $\tle{0\%}$ &         $5\%$ \\
\cline{3-19}
& & \multicolumn{2}{c|}{Multi-Neuron Relaxation} &  \t{k-ReLU}~\cite{singh2019beyond} &        $15\%$ &   $\mb{27\%}$ &         $2\%$ &        $23\%$ &   $\tle{0\%}$ &   $\tle{0\%}$ &   $\tle{0\%}$ &        $32\%$ &   $\tle{0\%}$ &   $\tle{0\%}$ &   $\tle{0\%}$ &   $\tle{0\%}$ &   $\tle{0\%}$ &   $\tle{0\%}$ \\
\cline{2-19}

& \multicolumn{3}{c|}{\multirow{2}{*}{SDP}} & \t{SDPVerify}~\cite{raghunathan2018semidefinite} &   $\tle{0\%}$ &   $\tle{0\%}$ &   $\tle{0\%}$ &   $\tle{0\%}$ &   $\tle{0\%}$ &   $\tle{0\%}$ &   $\tle{0\%}$ &   $\tle{0\%}$ &   $\tle{0\%}$ &   $\tle{0\%}$ &   $\tle{0\%}$ &   $\tle{0\%}$ &   $\tle{0\%}$ &   $\tle{0\%}$ \\

\cline{5-19}
 
& \multicolumn{3}{c|}{} & \t{LMIVerify}~\cite{fazlyab2019safety} &   $\tle{0\%}$ &   $\tle{0\%}$ &   $\tle{0\%}$ &   $\tle{0\%}$ &   $\tle{0\%}$ &   $\tle{0\%}$ &   $\tle{0\%}$ &   $\tle{0\%}$ &   $\tle{0\%}$ &   $\tle{0\%}$ &   $\tle{0\%}$ &   $\tle{0\%}$ &   $\tle{0\%}$ &   $\tle{0\%}$ \\
\cline{2-19}
& \multirow{3}{*}{Lipschitz} & \multicolumn{2}{c|}{\multirow{3}{*}{\shortstack{General\\ Lipschitz}}} & \t{Op-norm}~\cite{szegedy2013intriguing,tsuzuku2018lipschitz} &   $\tle{0\%}$ &   $\tle{0\%}$ &   $\tle{0\%}$ &   $\tle{0\%}$ &   $\tle{0\%}$ &   $\tle{0\%}$ &   $\tle{0\%}$ &   $\tle{0\%}$ &   $\tle{0\%}$ &   $\tle{0\%}$ &   $\tle{0\%}$ &   $\tle{0\%}$ &   $\tle{0\%}$ &   $\tle{0\%}$ \\
\cline{5-19}
 
& & \multicolumn{2}{c|}{} & \t{FastLip}~\cite{weng2018towards} &        $12\%$ &   $\mb{27\%}$ &   $\tle{0\%}$ &        $17\%$ &   $\tle{0\%}$ &        $17\%$ &   $\tle{0\%}$ &        $24\%$ &   $\tle{0\%}$ &   $\tle{0\%}$ &   $\tle{0\%}$ &   $\tle{0\%}$ &   $\tle{0\%}$ &   $\tle{0\%}$ \\
\cline{5-19}
& & \multicolumn{2}{c|}{} & \t{RecurJac}~\cite{zhang2019recurjac} &        $14\%$ &   $\mb{27\%}$ &         $2\%$ &        $17\%$ &   $\tle{0\%}$ &   $\tle{0\%}$ &   $\tle{0\%}$ &   $\tle{0\%}$ &   $\tle{0\%}$ &   $\tle{0\%}$ &   $\tle{0\%}$ &   $\tle{0\%}$ &   $\tle{0\%}$ &   $\tle{0\%}$ \\
\hline
\rowcolor{tabgray}
\multicolumn{5}{c|}{Accuracy under PGD (Upper Bound of Robust Accuracy)} &        $22\%$ &        $28\%$ &        $23\%$ &        $26\%$ &        $19\%$ &        $34\%$ &        $34\%$ &        $34\%$ &        $33\%$ &        $39\%$ &        $36\%$ &        $40\%$ &        $41\%$ &        $31\%$ \\
\hline
\rowcolor{tabgray}
\multicolumn{5}{c|}{Clean Accuracy} &        $33\%$ &        $31\%$ &        $37\%$ &        $30\%$ &        $26\%$ &        $39\%$ &        $44\%$ &        $46\%$ &        $53\%$ &        $48\%$ &        $52\%$ &        $46\%$ &        $66\%$ &        $46\%$ \\
\bottomrule

    \end{tabular}
    }
    \vspace{-1.0em}
    \label{tab:final_main_table}
\end{table*}

%% file: tables/final-main-table-2.tex
\begin{table*}[ht]
\caption{\emph{Certified accuracy} on CIFAR-10  certified by \emph{probabilistic approaches} for smoothed DNNs. The models are trained with different ``Robust Training" approaches and smoothed with distributions labelled as ``Smooth Dist.''. 
Numbers within each big cell under the rightmost column are comparable.
The \textbf{bolded} numbers mark the highest ones within each group.}
    \label{tab:exp-B-main-table}
    \centering
    \vspace{-1.0em}
    \resizebox{0.9\textwidth}{!}{
\begin{tabular}{c|l||l|l|l||cccccc}
\toprule
\multicolumn{1}{c|}{Adversary} & \multicolumn{1}{l||}{Model Structure} & \multicolumn{1}{l|}{Verification Approach} & \multicolumn{1}{l|}{Robust Training}           & \multicolumn{1}{l||}{Smooth Dist.} & \multicolumn{6}{l}{Certified Robust Accuracy under Perturbation Radius $\epsilon$}                               \\ 
\hline\hline
\multicolumn{5}{r}{$\epsilon=$} & $0.25$        & $0.50$         & $0.75$        & $1.00$         & $1.25$        & $1.50$         \\
\hline
\multirow{8}{*}{$\cL_2$}           & \multirow{3}{*}{Wide ResNet 40-2}    & Differential Privacy Based~\cite{lecuyer2019certified}                      & \multirow{3}{*}{Data Augmentation~\cite{cohen2019certified,yang2020randomized}}                      & \multirow{8}{*}{Gaussian}            & $34.2\%$      & $14.8\%$      & $6.8\%$       & $2.2\%$       & $0.0\%$       & $0.0\%$       \\
\cline{3-3}
                               &                                                                    & Neyman-Pearson~\cite{cohen2019certified,teng2020ell,yang2020randomized,Zhang2020BlackBoxCW}                           &                                                         &                                      & $\mb{68.8\%}$ & $\mb{46.8\%}$ & $\mb{36.0\%}$ & $\mb{25.4\%}$ & $\mb{19.8\%}$ & $\mb{15.6\%}$ \\
\cline{3-3}
                               &                                                                  & $f$-Divergence~\cite{Dvijotham2020A}                             &                                                         &                                      & $62.2\%$      & $41.8\%$      & $27.2\%$      & $19.2\%$      & $14.2\%$      & $11.4\%$      \\
\cline{2-4}\cline{6-11}
                               &                              \multirow{5}{*}{ResNet-110}          & \multirow{5}{*}{Neyman-Pearson~\cite{cohen2019certified,teng2020ell,yang2020randomized,Zhang2020BlackBoxCW}}            & Data Augmentation~\cite{cohen2019certified,yang2020randomized}                                       &                                      & $61.2\%$      & $43.2\%$      & $32.0\%$      & $22.4\%$      & $17.2\%$      & $14.0\%$      \\
\cline{4-4}
                               &                                                                   &                                            & Adversarial Training~\cite{salman2019provably}                                    &                                      & $73.0\%$      & $57.8\%$      & $48.2\%$      & $\mb{37.2\%}$ & $33.6\%$      & $\mb{28.2\%}$ \\
\cline{4-4}
                              &                                                                   &                                            & Adversarial + Pretraining~\cite{salman2019provably,carmon2019unlabeled}                               &                                      & $\mb{81.8\%}$ & $\mb{62.6\%}$ & $\mb{52.4\%}$ & $37.2\%$      & $\mb{34.0\%}$ & $30.2\%$      \\
\cline{4-4}
                               &                                                                   &                                            & \t{MACER}~\cite{Zhai2020MACER:}                                                   &                                      & $68.8\%$      & $52.6\%$      & $40.4\%$      & $33.0\%$      & $27.8\%$      & $25.0\%$      \\
\cline{4-4}                               &                                                                   &                                            & \t{ADRE}~\cite{feng2020regularized}                                                    &                                      & $68.0\%$      & $50.2\%$      & $37.8\%$      & $30.2\%$      & $23.0\%$      & $17.0\%$      \\
\hline \hline
\multicolumn{5}{r}{$\epsilon=$}  & $0.5$         & $1.0$         & $1.5$         & $2.0$         & $3.0$         & $4.0$         \\
\hline
 \multirow{4}{*}{$\cL_1$}         & \multirow{4}{*}{Wide ResNet 40-2}    & Differential Privacy Based~\cite{lecuyer2019certified}                      & \multirow{4}{*}{Data Augmentation~\cite{cohen2019certified,yang2020randomized}}                      & \multirow{3}{*}{Laplace}             & $43.0\%$      & $20.8\%$      & $12.2\%$      & $7.2\%$       & $1.4\%$       & $0.0\%$       \\
\cline{3-3}
                               &                                                                  & R\'enyi Divergence~\cite{li2019certified}          &                                                         &                                      & $58.2\%$      & $39.4\%$      & $27.0\%$      & $16.8\%$      & $9.2\%$       & $4.0\%$       \\
\cline{3-3}
                              &                                                                   & \multirow{2}{*}{Neyman-Pearson~\cite{cohen2019certified,teng2020ell,yang2020randomized,Zhang2020BlackBoxCW}}            &                                                         &                                      & $58.4\%$      & $39.6\%$      & $27.0\%$      & $17.2\%$      & $9.2\%$       & $4.2\%$       \\
\cline{5-5}
                               &                                                                  &                                            &                                                         & Uniform                              & $\mb{69.2\%}$ & $\mb{56.6\%}$ & $\mb{48.0\%}$ & $\mb{39.4\%}$ & $\mb{26.0\%}$ & $\mb{20.4\%}$ \\
\hline \hline
\multicolumn{5}{r}{$\epsilon=$}   & $1/255$       & $2/255$       & $4/255$       & $8/255$       &               &               \\
\hline
\multirow{2}{*}{$\cL_\infty$}     & \multirow{2}{*}{Wide ResNet 40-2}    & \multirow{2}{*}{Neyman-Pearson~\cite{cohen2019certified,teng2020ell,yang2020randomized,Zhang2020BlackBoxCW}}            & Data Augmentation~\cite{cohen2019certified,yang2020randomized}                                       & \multirow{2}{*}{Gaussian}            & $71.4\%$      & $52.0\%$      & $29.0\%$      & $12.8\%$      &               &               \\
\cline{4-4}
                              &                                         &              & Adversarial Training~\cite{salman2019provably}          &      & $\mb{83.2\%}$      & $\mb{65.0\%}$      & $\mb{49.6\%}$      & $\mb{25.4\%}$      &               &   \\
\bottomrule
\end{tabular}
}
\vspace{-1.85em}
\end{table*}